%% file: car_usage_classification_kdd.tex
\documentclass[sigconf]{acmart}

\usepackage{booktabs} 
\usepackage{color}
\usepackage{subcaption}
\usepackage{hyphenat}
\usepackage[linesnumbered,ruled]{algorithm2e}
\usepackage{pbox}

\setcopyright{none}

\acmDOI{10.475/123_4}

\acmISBN{123-4567-24-567/08/06}

\acmPrice{15.00}

\begin{document}
\title{Ridesourcing Car Detection by Transfer Learning}

\author{Leye Wang$^1$, Xu Geng$^1$, Jintao Ke$^1$, Chen Peng$^1$, Xiaojuan Ma$^1$, Daqing Zhang$^2$, Qiang Yang$^{1}$}
\affiliation{%
  \institution{$^1$Hong Kong University of Science and Technology}
  \institution{$^2$Key Lab of High Confidence Software Technologies, Peking University}
  \institution{wly@cse.ust.hk, \{xgeng, jke, cpengac\}@connect.ust.hk, mxj@cse.ust.hk, dqzhang@sei.pku.edu.cn, qyang@cse.ust.hk}
}







\renewcommand{\shortauthors}{L. Wang et al.}

\begin{abstract}
Ridesourcing platforms like Uber and Didi are getting more and more popular around the world. However, unauthorized ridesourcing activities taking advantages of the sharing economy can greatly impair the healthy development of this emerging industry. As the first step to regulate on-demand ride services and eliminate black market, we design a method to detect ridesourcing cars from a pool of cars based on their trajectories. Since licensed ridesourcing car traces are not openly available and may be completely missing in some cities due to legal issues, we turn to transferring knowledge from public transport open data, i.e, taxis and buses, to ridesourcing detection among ordinary vehicles. We propose a two-stage transfer learning framework. In Stage~1, we take taxi and bus data as input to learn a random forest (RF) classifier using trajectory features shared by taxis/buses and ridesourcing/other cars. Then, we use the RF to label all the candidate cars. In Stage~2, leveraging the subset of high confident labels from the previous stage as input, we further learn a convolutional neural network (CNN) classifier for ridesourcing detection, and iteratively refine RF and CNN, as well as the feature set, via a co-training process. Finally, we use the resulting ensemble of RF and CNN to identify the ridesourcing cars in the candidate pool. Experiments on real car, taxi and bus traces show that our transfer learning framework, with no need of a pre-labeled ridesourcing dataset, can achieve similar accuracy as the supervised learning methods.
\end{abstract}

%




\maketitle

\input{introduction}
\input{problem}
\input{overview}
\input{solution}
\input{evaluation}
\input{related_work}
\input{conclusion}
\bibliographystyle{ACM-Reference-Format}
\bibliography{car_usage_classification_ref} 

\end{document}

%% file: introduction.tex
\section{Introduction} 
\label{sec:introduction}

With the prevalence of online ridesourcing~\cite{sharedMobility} platforms like Uber~\cite{Uber}, Lyft~\cite{Lyft} and Didi Express~\cite{Didi}, private cars now can offer on-demand ride services to passengers quite easily. While enhancing urban mobility, ridesourcing services have a variety of social-economic issues, especially if accompanied by black market activities. First of all, passengers may be exposed to safety and financial risks. For example, some ridesourcing drivers may trade customer orders with an unauthorized driver who might not have proper training or sufficient driving experience, and might not provide passenger with insurance coverage~\cite{DIdiOrderHacked}. The passenger might get ripped off, or even worse, encounter severe crimes like sexual harassment, robbery, etc.\footnote{A list of incidents about ridesourcing can be found in  \url{http://www.whosdrivingyou.org/rideshare-incidents}. More specifically, the incidents incurred by unauthorized drivers can be found in the category of `\textit{Imposters}'. (Accessed: 2016-02-09)}
For example, in May 2016, the driver of an unauthorized ridesourcing car with a fake numberplate robbed and killed a passenger in China~\cite{DidiKill}. Second, the safety of unlicensed drivers may also be at risk, as previous report suggested that `gypsy cab' drivers are also easy targets of assaults~\cite{Dangerous_gypsy_cab}. Third, ridesourcing companies may lose both profits and credibility if their licensed drivers bypass the platforms and make under-the-table deals with customers~\cite{uber_under_table}. Fourth, other businesses have little information of whether their employees illegally use the company-owned cars for ridesourcing, which brings in potential management and economic risks to the employers~\cite{Company_car_didi}. In summary, despite the environmental and social benefits introduced by this rapidly-growing business, the existence of black market and unregulated activities can seriously impair the credibility and healthy development of the whole ridesourcing industry~\cite{mcbride2015ridesourcing}.

If we could detect unauthorized ridesourcing cars, we would be able to take further measures to alleviate the above-mentioned issues to a large extent. 
In this paper, as the first step towards this direction, we aim to address the problem of detecting ridesourcing cars from a pool of vehicles,
assuming that the recent trajectories of all the candidate vehicles are accessible. This assumption can be technically realized. For instance, with the huge number of street/red light cameras~\cite{NYC_camera,norris1999maximum} and mature car numberplate recognition techniques~\cite{ondrej2007algorithmic}, it is possible to reconstruct the moving traces of most vehicles in the urban area~\cite{hunter2014path}.  In addition, many companies deploy position tracking devices on their corporate vehicles to prevent the abuse by employees.

An intuitive solution to this research problem is directly comparing the traces of the candidate cars with ride trajectory obtained from ridesourcing platforms (uploaded from drivers' smartphones during ridesourcing trips). However, this approach suffers from several pitfalls. First, access to the recent trace data stored in various ridesourcing platforms is probably difficult due to concerns like user privacy~\cite{yang2011information}. Second, ridesourcing services are still regarded as illegal in many places around the world~\cite{wiki_uber_legal}. It means that no ridesourcing platform would have data for these places; however, the need for regulating the ridesourcing black market may be even greater in such places~\cite{austin_black_market}.

With the objective of proposing an easy-to-deploy and comprehensive ridesourcing car detection algorithm, we turn to \textit{taxi} and \textit{bus} traces for \textit{knowledge transfer}  to avoid the aforementioned pitfalls. For one thing, taxi and bus trajectories are perhaps the most openly available types of vehicle trace data~\cite{beijingTaxi,NYTaxi,SFTaxi,romaTaxi,shanghaiTaxi,ChicagoTaxi,Irish_bus_data,rice-ad_hoc_city-20030911}. For another, previous study has shown that taxis and ridesourcing cars share many mobility similarities~\cite{chen2017understanding}, indicating the transfer feasibility; while bus traces can serve as negative cases, as buses are distinct from ridesourcing cars in moving patterns. In spite of the feasibility of the knowledge transfer idea, we still face some issues. For instance, taxi data, even the most frequently updated ones~\cite{ChicagoTaxi,NYTaxi}, are published with a substantial delay due to data cleaning and pre-processing. Additionally, more taxi open data only cover a specific period~\cite{shanghaiTaxi,beijingTaxi,SFTaxi,romaTaxi}. In other words, the time span of taxi data may not align with the time span of the candidate rides, especially if we want to use the most recent driving patterns of the candidate cars for ridesourcing detection. 

In this paper, we design a two-stage transfer learning framework for adapting taxi and bus knowledge to ridesourcing detection. 
In \textbf{Stage~1}, we extract the features shared by taxis and ridesourcing cars, and also durable to the time misalignment issue above-mentioned. Using these features, we train a random forest (RF)~\cite{liaw2002classification} classifier with taxi and bus open data, and adopt the classifier to predict whether a candidate car is ridesourcing or not, denoted as \textit{s1-label}. However, the features extracted from taxi open data, although effective in ridesourcing detection to a certain extent, may still not fully characterize ridesourcing cars. In \textbf{Stage~2}, we attempt to find more ridesourcing-specific features to improve classification accuracy. To this end, instead of leveraging all s1-labels, we build a training set consisting only of the cars whose s1-labels are predicted with \textit{high confidence}, and reset all the other cars as unlabeled.  With this training set as input, we start an iterative \textit{co-training} process~\cite{blum1998combining,goldman2000enhancing} to learn ridesourcing-specific features from the remaining unlabeled candidate car data. In this stage, besides RF, we design a convolutional neural network (CNN)~\cite{krizhevsky2012imagenet} classifier to enrich ridesourcing-specific features for detection. In each iteration, we update both RF and CNN on the current training set, use each classifier to predict the labels of the remaining cars, and add the newly confidently classified cars into the training set. The co-training process stops when no new cars are added to the training set. Afterwards, we use the ensemble of RF and CNN to determine the final label of each candidate car.

Briefly, this paper makes the following contributions:

(1) To the best of our knowledge, this is the first work on ridesourcing car detection. In particular, by properly transferring knowledge from taxi and bus open data, no labeled ridesourcing dataset is needed in our proposed method.

(2) We propose a two-stage framework to transfer taxi and bus knowledge for ridesourcing car detection. The first stage, with taxi and bus open data as input, identifies ridesourcing cars using features that are shared by both taxis and ridesourcing cars and can withstand the open data limitation like time misalignment. Taken the cars classified with high confidence in the first stage as input, the second stage leverages a co-training process to further infer ridesourcing-specific features. Two classifiers, RF and CNN, are iteratively refined, and the final ridesourcing label for each candidate car is determined by an ensemble of the two classifiers.

(3) Experiment results on the traces of about $10,000$ cars over seven workdays, among which $600$ cars are manually labeled by majority voting as test data, have shown that our transfer learning method, \textit{with no need of the labeled ridesourcing dataset}, can achieve an overall detection accuracy of $85\%$, which is comparable to supervised learning methods, as well as the average manual label accuracy (considering majority voting as ground truth).


%% file: problem.tex
\section{Problem Formulation} 
\label{sec:problem_formulation}

In this section, we first formulate our research problem from the application perspective. Then, we abstract our problem in the transfer learning settings~\cite{pan2010survey}.

\textbf{Ridesourcing Car Detection Problem.} Given a set of $n$ cars' traces $\{ \mathcal T_1, \mathcal T_2, \cdots ,\mathcal T_n \}$, where the $i$th car's is $\mathcal T_i = \langle p_1, p_2, \cdots, p_m \rangle$ and $p_j = \langle lat_j, lon_j, time_j \rangle$. $lat_j$, $lon_j$ and $time_j$ are the latitude, longitude and timestamp of the $j$th trace point. We aim to identify the cars undertaking ridesourcing activities from $n$ candidate vehicles.\footnote{We suppose no candidate car is taxi, as taxis can be easily filtered out because their information is officially registered in the government. }

Note that in reality people can choose to be part-time ridesourcing drivers (e.g., one day per week or even per month) and fine differentiation can be performed based on the frequency of service. In this paper, however, we focus on identifying \textit{routine ridesourcing} cars, i.e., frequently performing ridesourcing services (e.g., in most workdays), and thus we formulate the above problem as binary classification. For brevity, in the rest of the papers, we use \textit{ridesourcing} cars to refer to \textit{routine ridesourcing} cars. Our future work will consider a more fine-grained categorization, e.g., differentiating between part-time and full-time ridesourcing cars.


Binary classification is an application-level problem formulation. As we propose to address the problem by transferring knowledge from taxi and bus open data, we further formulate the problem mathematically in the transfer learning aspect.

\textbf{Transfer Learning Formulation}. The source domain in our problem is the traces of public transport vehicles, namely, taxis and buses in this paper; each instance is labeled to indicate whether it is a taxi or a bus. 
\begin{align}
	&X_s = \{x_{si}\}, \quad Y_s= \{y_{si}\} \\
	&X_{st} = \{x_{si}|y_{si}=\textit{`taxi'}\}, \quad X_{sb} = \{x_{si}|y_{si}=\textit{`bus'}\}
\end{align}
where $x_{si}$ is a vehicle instance in the source domain $X_s$; $X_{st}$ and $X_{sb}$ denote the set of taxis and buses in the source domain, respectively.

The target domain includes the traces of candidate cars for ridesourcing detection; each instance corresponding to one candidate car with no initial labels: 
\begin{equation}
	X_t = \{x_{ti}\}
\end{equation}

Our objective is to learn a classification function in the target domain for ridesourcing car detection, $\mathcal F: X_t \mapsto \{\textit{`ridesourcing'},\textit{`other'}\}$, aiming to maximize:
\begin{align}
	& \textstyle \arg \max_{_\mathcal F}\ Pr(X_{tr}^{\mathcal F}, X_{to}^{\mathcal F}|X_t) \\
	\text{where} \quad & X_{tr}^{\mathcal F} = \{X_{ti}| \mathcal F(X_{ti}) = \textit{`ridesourcing'} \} \\
	& X_{to}^{\mathcal F} = \{X_{ti}| \mathcal F(X_{ti}) = \textit{`other'} \}
\end{align}

According to~\cite{pan2010survey}, traditional transfer learning problems on classification usually have either some labeled data in the target domain (inductive transfer learning) or the same tasks in both domains (transductive transfer learning), neither of which applies to our problem. On one hand, we have no labels in the target domain. On the other hand, our task in the target domain $X_t \mapsto \{\textit{`ridesourcing'},\textit{`other'}\}$ is not exactly same as, although related to, the task in the source domain $X_s \mapsto \{\textit{`taxi'},\textit{`bus'}\}$.  This brings new challenges to our problem and requires a novel solution. 

Actually, although the two tasks in our source and target domains are not identical, somehow, the classes of the source domain, taxis/buses, can be regarded as special cases of the classes of the target domain, ridesourcing/other. In other words, if performing like taxis, the cars in the target domain probably undertake ridesourcing activities; if performing like buses, they would not be ridesourcing cars. This ensures the feasibility of transfer learning and the next section will elaborate our method in detail.


%% file: overview.tex

\section{Methodology} 
\label{sec:method_overview}

In this section, we propose a two-stage learning framework to address our research problem. 

\subsection{Basic Idea} 
\label{sub:basic_idea}
\begin{figure}[t]
	\centering
	\includegraphics[width=1\linewidth]{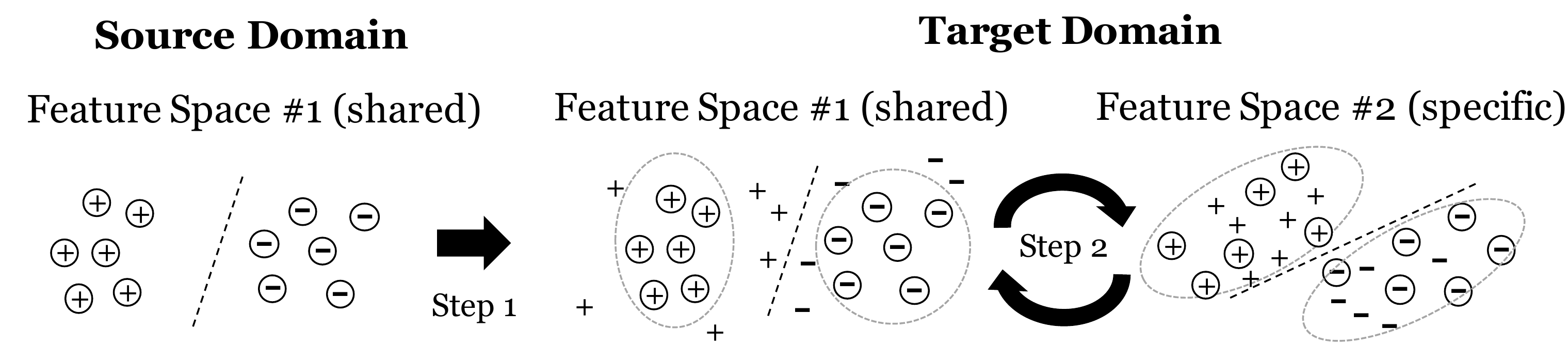}
	\vspace{-2em}
	\caption{Basic idea of our transfer learning solution. (The dashed circles are high-confident classification boundary.)}
	\label{fig:basic_idea}
	\vspace{-1em}
\end{figure}

 In our problem, although tasks are not exactly same between the source and target domains, each category in the source domain could be seen as a special case of a corresponding category in the target domain --- if an unlabeled vehicle moves very similar to taxis, it is probably a ridesourcing car; if its mobility pattern is close to buses, it should not undertake ridesourcing activities.\footnote{The inverse statement may not be true, e.g., a non-ridesourcing car does not have to act like buses.} Then, if a proper feature space shared by our source and target domains is found, we could expect to use the patterns learned from the source domain to classify the unlabeled cars in the target domain. The step 1 in Figure~\ref{fig:basic_idea} illustrates this process. 

Note that, as special cases, such classified cars in the target domain may be biased, i.e., gathering closely in the shared feature space, and not fully characterize the whole set of ridesourcing/other cars. Then, to find more ridesourcing cars out of this special scope, we try to find another feature space to re-map the already classified cars. Previous studies have shown that, as long as the two feature spaces are independent of each other, it is probable that the instances gathering in one space will be scattered in the other~\cite{blum1998combining}. Therefore, if we can find such a second feature space, we could detect more ridesourcing/other cars which are not recognized in the first feature space; with the newly classified cars in the second space, returning to the first space, we can find more confident ridesourcing/other cars. This iterative process for refining the classifier from two distinct feature spaces are called \textit{co-training}~\cite{blum1998combining}. This technique helps us to jump out of the special cases learned from the source domain, and thus we can build a classifier that is able to detect ridesourcing cars more comprehensively.

To make this basic idea work effectively, the two feature spaces must be carefully modeled. On one hand, the features in the first space need to be shared by the taxi/bus instances in the source domain and the ridesourcing/other instances in the target domain, especially considering the time misalignment issue (the time span of taxi/bus traces in the source domain may not be the same as the car traces in the target domain). If the feature space is not shared well, the learned patterns from the source domain will not be useful in the target domain. On the other hand, the features in the second space should be distinct from those in the first space as much as possible, in order to make the co-training process efficient. As the second feature space is only used in the target domain, it can focus more on the ridesourcing-specific features.  

\begin{figure}[t]
	\centering
	\includegraphics[width=1\linewidth]{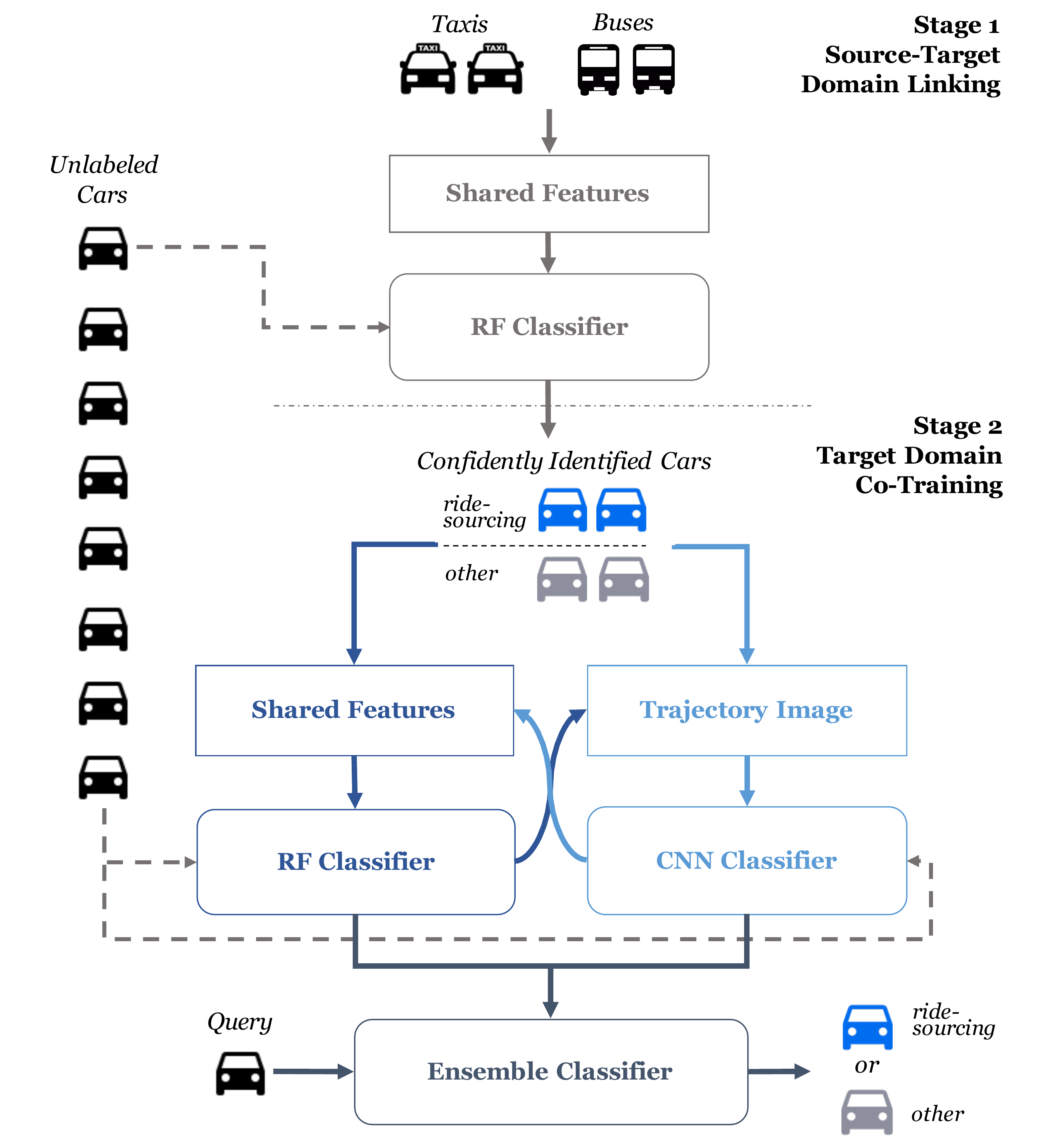}
	\vspace{-2em}
	\caption{Framework overview.}
	\label{fig:overview}
	\vspace{-1.5em}
\end{figure}

\subsection{Framework Overview} 
\label{sub:overview}

Following the basic idea, we design a two-stage learning framework to detect ridesourcing cars leveraging knowledge transferred from taxi and bus open traces. Figure~\ref{fig:overview} is an overview of our framework. The two stages are called as \textit{source-target domain linking} and \textit{target domain co-training}, respectively.

\textit{Source-Target Domain Linking}. In Stage~1, based on taxi and bus data in the source domain, we attempt to detect cars in the target domain which are similar to taxi/bus, so that we can confidently classify them to ridesourcing/other. To this end, we need to select the features that are shared by taxis and ridesourcing cars. In addition, the selected features should be durable to the limitations  of using public taxi data, such as time misalignment. With these considerations in mind, we mainly extract features from distance and coverage aspects and then learn a random forest (RF)~\cite{liaw2002classification} based on source domain data. We then apply RF to the target domain unlabeled cars and keep the predicted labels of the cars which are classified with high confidence.

\textit{Target Domain Co-training}. Using the traces of cars confidently labeled as ridesourcing/other in Stage~1 as the initial training set, Stage~2 leverages the co-training technique~\cite{blum1998combining} to extract more ridesourcing-specific features for comprehensive ridesourcing detection. The basic idea of co-training is to iteratively refine two classifiers by adding new confidently classified instances into the training set; the more distinct the two classifiers are from each other, the better co-training performance is~\cite{blum1998combining}. Hence we design a convolutional neural network (CNN)~\cite{zhang2016deep,krizhevsky2012imagenet} as the second classifier,  the input of which is the trajectory image generated from a car's daily GPS traces. Different from the input features used in the first classifier RF, a trajectory image of a car is constructed by splitting the city area into $M \times N$ grids, i.e., pixels, and setting each pixel to a lighter (darker) color if the car stays in the corresponding city grid for a longer (shorter) period of time.

We elaborate on the details of each stage in the next subsection.


%% file: solution.tex

\subsection{Source-Target Domain Linking} 
\label{sub:taxi_knowledge_transfer}

To use taxi knowledge for ridesourcing car detection, we first study which features of taxi would be likely to exist also for ridesourcing cars, i.e., \textit{shared features}. Then, based on such features, we build a random forest classifier for detecting ridesourcing cars.

\vspace{+.5em}
\textbf{Shared Features}

Finding shared features, in other words, is to identify the features that both taxis and ridesourcing cars share similar values, while the other cars have dissimilar values. This is a tricky issue, because although taxis share certain trajectory patterns with ridesourcing cars, they also differ in some aspects~\cite{chen2017understanding}. Besides, while taxi data are relatively open, the time span of taxi data and unlabeled car data may not be aligned. With these issues in mind, we identify the following shared features.

\textit{Distance}. Driving distance is intuitively an effective measurement for detecting taxi and ridesourcing cars, as they would be likely to run much longer distance than most of the other cars. We thus extract cars' daily mean driving distance and its variance as shared features. It is also worth noting that, previous studies have pointed out that compared to taxis' 24-hour work style, ridesourcing cars rarely work after 00:00 and before 6:00~\cite{chen2017understanding}. Therefore, when calculating the distance in taxi data we only take the time span of 6:00 to 24:00 into consideration.

\begin{figure}[t]
    \centering
    \begin{subfigure}[t]{0.22\linewidth}
        \includegraphics[width=1\linewidth]{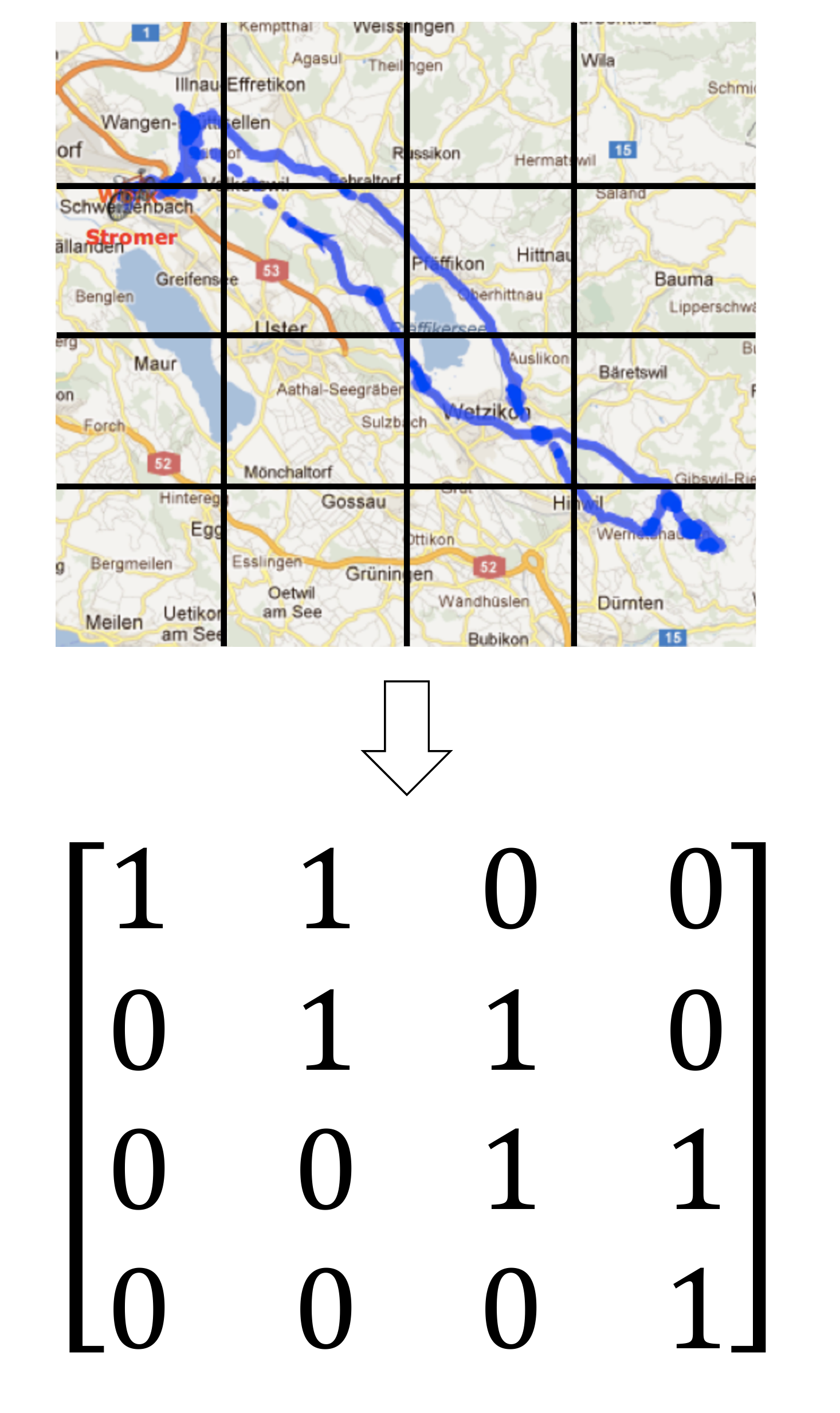}
        \caption{Conversion}
        \label{fig:cov_matrix}
    \end{subfigure}
    \hspace{.7cm}
    \begin{subfigure}[t]{0.42\linewidth}
        \includegraphics[width=1\linewidth]{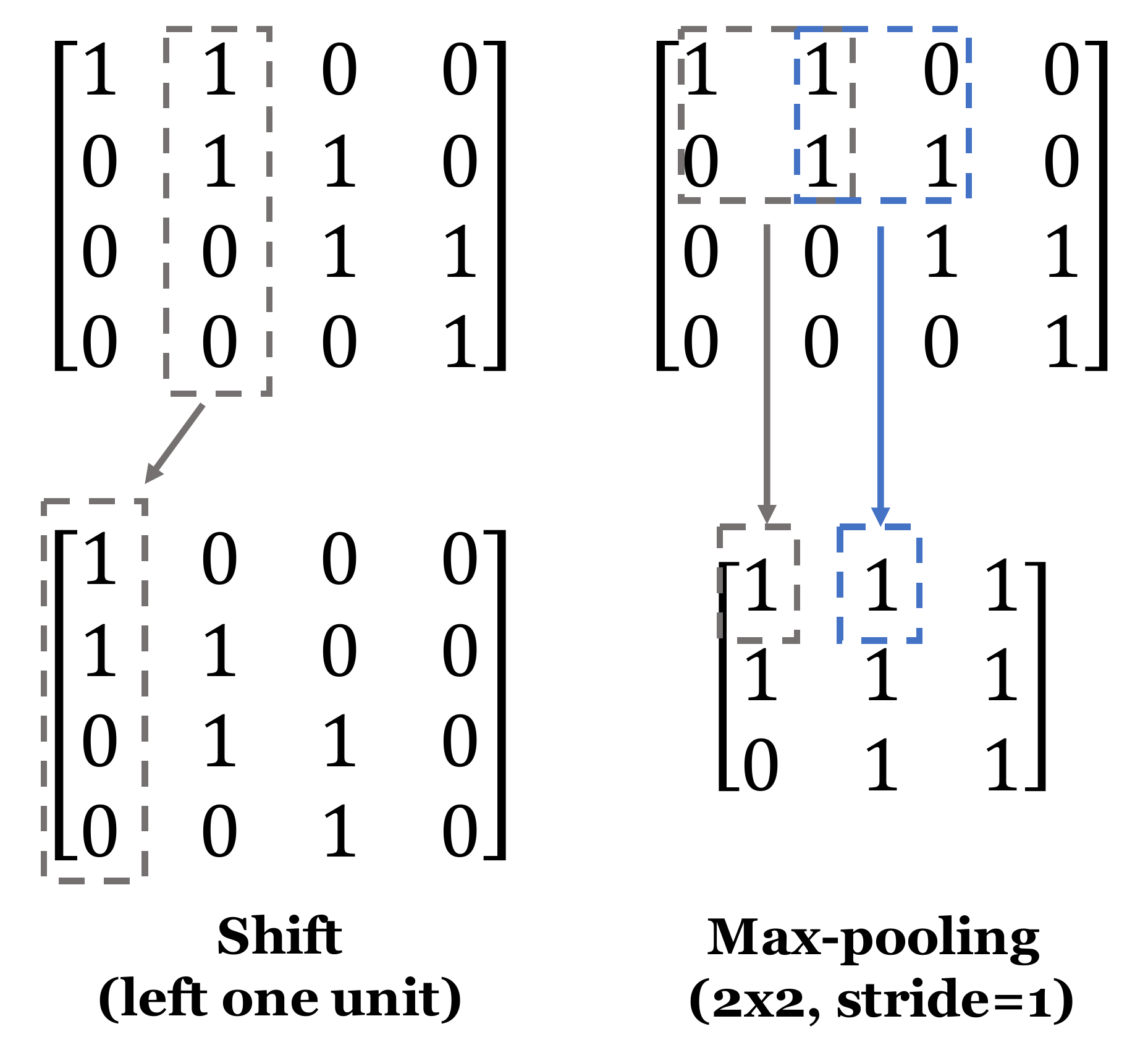}
        \caption{Operations}
        \label{fig:cov_matrix_op}
    \end{subfigure}
    \vspace{-1em}
    \caption{Coverage matrix example.}
    \vspace{-2em}
\end{figure}

\textit{Coverage}. Coverage is another important aspect that could be shared between taxis and ridesourcing cars. Taxis and ridesourcing cars' origins and destinations are determined by their served customers and the trips are usually short and scattered, hence the city area covered is probably larger than most of the other cars (e.g., family usage). To get coverage-related features, we first split the city area into $M \times N$ grids, and then mark a grid as $1$ if it is visited by a car during a certain time slot; otherwise $0$. To distinguish rush hours in the morning and afternoon, we split one day into three time slots: 6:00 to 12:00, 12:00 to 18:00, and 18:00 to 24:00; the time span of 00:00 to 6:00 is discarded like the distance feature. We also consider a whole time span, i.e., 6:00 to 24:00 as another time slot to catch daily patterns. Figure~\ref{fig:cov_matrix} shows an example of converting a car's trace to a coverage matrix.

Denote this 1-0 binary coverage matrix as $C_k^z$ for the $z$th time slot of the $k$th day (totally $n$ days), we can calculate the \textit{daily mean coverage count for each time slot} and its variance as features:
\setlength{\belowdisplayskip}{1em} \setlength{\belowdisplayshortskip}{1pt}
\setlength{\abovedisplayskip}{1em} \setlength{\abovedisplayshortskip}{1pt}
\begin{align}
    &\textit{mean coverage: } & \textstyle\sum_{i,j,k} C_k^z[i,j] / n , & \qquad z = 0, 1, 2, 3  \\
    &\textit{coverage variance: } & \mathit{var}(\{\textstyle\sum_{i,j} C_k^z[i,j]\}), & \qquad z = 0,1, 2, 3
\end{align}
where $0$th time slot represents the whole time span of 6:00 to 24:00.

\textit{Robust Coverage Similarity Metric}. Besides mean coverage count, we also use features to quantify the variation between different coverage matrices (intra- and inter-day). An intuitive way is directly using cell-to-cell similarity metrics like Jaccard. However, the possible GPS error may map a taxi's emergence in cell $C[i,j]$ to another (nearby) cell $C[i',j']$, making cell-to-cell comparison noisy. Besides, intuitively, the similarity of two matrices covering nearby cells should be higher than that of two matrices covering faraway cells, e.g.,
\begin{equation*}
\footnotesize
\setlength{\belowdisplayskip}{1em} \setlength{\belowdisplayshortskip}{0pt}
\setlength{\abovedisplayskip}{1em} \setlength{\abovedisplayshortskip}{0pt}
    sim(
    \begin{bmatrix}
    1 & 0 & 0\\
    0 & 0 & 0\\
    0 & 0 & 0
    \end{bmatrix}, 
    \begin{bmatrix}
    0 & 1 & 0 \\
    0 & 0 & 0 \\
    0 & 0 & 0
    \end{bmatrix})
    >
    sim(
    \begin{bmatrix}
    1 & 0 & 0\\
    0 & 0 & 0\\
    0 & 0 & 0
    \end{bmatrix}, 
    \begin{bmatrix}
    0 & 0 & 0 \\
    0 & 0 & 0 \\
    0 & 0 & 1
    \end{bmatrix})
\end{equation*}
But the normal cell-to-cell metric like Jaccard cannot take this into account. To address these issues, we compute a robust coverage matrix similarity metric, as shown in Algorithm~\ref{alg:coverage_diff}. 

\setlength{\textfloatsep}{3pt}
\begin{algorithm}[t]
\footnotesize
    \SetKwInOut{Input}{Input}
    \SetKwInOut{Output}{Output}
    \Input{$C_1, C_2$: two coverage matrices.}
    \Output{$\mathit{sim}^*$: robust similarity measurement between $C_1$ and $C_2$.}
    \tcc{similarity on raw matrices}
    $[C_{l}, C_{r}, C_{u}, C_{d} ]= \mathit{matrix\_shift} (C_1)$; \tcc{shift 4 directions}
    $\mathbb{C}_1 = [C_1, C_{l}, C_{r}, C_{u}, C_{d} ]$ \;
    $\mathit{sim}_1$ = $\mathit{mean}(\mathit{Jaccard\_sim}(C_{i},C_2)\ \mathbf{for} \ C_{i}\ \mathbf{in}\ \mathbb{C}_1 )  $\;
   

    \tcc{similarity on max-pooling matrices}
    $C_1^{\mathit{p}} = \mathit{max\_pooling}(C_1)$ \;
    $C_2^{\mathit{p}} = \mathit{max\_pooling}(C_2)$ \;
    $[C_{l}^{\mathit{p}}, C_{r}^{\mathit{p}}, C_{u}^{\mathit{p}}, C_{d}^{\mathit{p}} ]= \mathit{matrix\_shift} (C_1^{\mathit{p}})$ \;
    $\mathbb{C}_1^{\mathit{p}} = [C_1^{\mathit{p}}, C_{l}^{\mathit{p}}, C_{r}^{\mathit{p}}, C_{u}^{\mathit{p}}, C_{d}^{\mathit{p}} ]$ \;
    $\mathit{sim}_2$ = $\mathit{mean}(\mathit{Jaccard\_sim}(C_{i},C_2^{\mathit{p}})\ \mathbf{for} \ C_{i}\ \mathbf{in}\ \mathbb{C}_1^{\mathit{p}} )  $\;
    $\mathit{sim}^* = \mathit{mean}(\mathit{sim_1},\mathit{sim_2})$ \;
    \Return $\mathit{sim}^*$ \;
    \caption{Robust Coverage Matrix Similarity}
    \label{alg:coverage_diff}
\end{algorithm}

Our intuition in the robust similarity metric is measuring the traditional cell-to-cell similarity metric between not only two raw matrices, but also the matrices operated by the shift and max-pooling functions~\cite{krizhevsky2012imagenet}, as shown in Figure~\ref{fig:cov_matrix_op}. Shift and max-pooling functions enable that when we calculate the similarity, a cell $C_1[i,j]$ is compared to not only $C_2[i,j]$ but also its nearby cells.

With this coverage similarity metric, we now measure both the \textit{intra-} and \textit{inter-day} coverage similarities as follows (totally $n$ days):
\setlength{\belowdisplayskip}{.5em} \setlength{\belowdisplayshortskip}{0pt}
\setlength{\abovedisplayskip}{.5em} \setlength{\abovedisplayshortskip}{0pt}
\begin{align}
    \textit{intra-day: } & \textstyle\frac{1}{n}  \textstyle\sum_{k} \frac{\mathit{sim}^*(C_k^1, C_k^2)+\mathit{sim}^*(C_k^1, C_k^3)+\mathit{sim}^*(C_k^2, C_k^3)}{3} \\
    \textit{inter-day: } & \textstyle\frac{2}{n(n-1)}  \textstyle\sum_{k'<k''} \mathit{sim}^*(C_{k'}^z, C_{k''}^z), \qquad z = 0,1,2,3
\end{align}

In summary, our extracted shared features are generally high-level statistics. Hence, even time misalignment issue exists, such shared features could be durable. For example, as time goes, new hotspots like commercial centers and residential areas may emerge in the city; then, the city cell stay-time distribution of taxis may be biased obviously to the new hotspots. However, our selected shared features, such as daily driving distance and number of covered cells, should not change significantly, because a taxi's working time is always limited to 24 hours per day.\footnote{Actually, the CNN built in Stage~2 just uses the cell stay-time as feature (modeling as image pixel color, see Sec.~\ref{sub:co_training}). In the experiment, we will verify that CNN performs much worse than RF if adopted in Stage~1, perhaps due to the time misalignment issue.}

\vspace{+.5em}
\textbf{Random Forest Classifier}

While taxi data can serve as `positive cases' similar to ridesourcing cars, to train a binary classifier, we also introduce \textit{bus} data as `negative cases' dissimilar to ridesourcing cars. 

Two major reasons exist for using bus data as negative cases. First, buses mostly follow regular trips. This characteristics is similar to family-usage cars, in which the most common workday travels are commuting. Hence, bus data are expected to help identify such family-usage cars. Second, like taxi data, bus data are often open~\cite{rice-ad_hoc_city-20030911,shanghaiTaxi,Irish_bus_data}. Even when the raw bus trace cannot be obtained, we can accurately simulate it according to the bus route and timetable information. Note that as cars for commuting mostly have two trips in one workday, thus for one bus instance, we only use two trip trajectories, one in the morning rush hours while the other in the evening rush hours, to extract features.

Using both taxi and bus data based on the shared features, we build a random forest (RF) classifier~\cite{liaw2002classification}. Applying this classifier on the target domain unlabeled cars, we can learn each car's label and its confidence. We then keep the labels of the cars which are classified with high confidence as the input to the next stage.


\subsection{Target Domain Co-Training} 
\label{sub:co_training}

With the high-confidently labeled cars from Stage~1, we leverage the co-training~\cite{blum1998combining} technique to discover more ridesourcing-specific features from the target domain to improve detection accuracy. More specifically, we construct the second classifier, a convolutional neural network (CNN) with input of trajectory images mapped from car traces. In this section, we will first illustrate CNN, and then summarize the overall co-training process.


\vspace{+.5em}
\textbf{Convolutional Neural Network Classifier} 

Recently, CNN begins to be leveraged in the spatio-temporal transportation data mining, such as traffic flow prediction~\cite{zhang2016deep}. The basic idea of such studies is to first create a city image where each pixel (e.g., 1km $\times$ 1km grid) represents the concerned transportation information (e.g., traffic inflow or outflow~\cite{zhang2016deep}). Inspired by this idea, we build a CNN for ridesourcing car detection with a car's gray-scale \textit{trajectory images} as input. 

\begin{figure*}[t]
    \centering
    \begin{subfigure}[t]{0.3\linewidth}
        \includegraphics[width=1\linewidth]{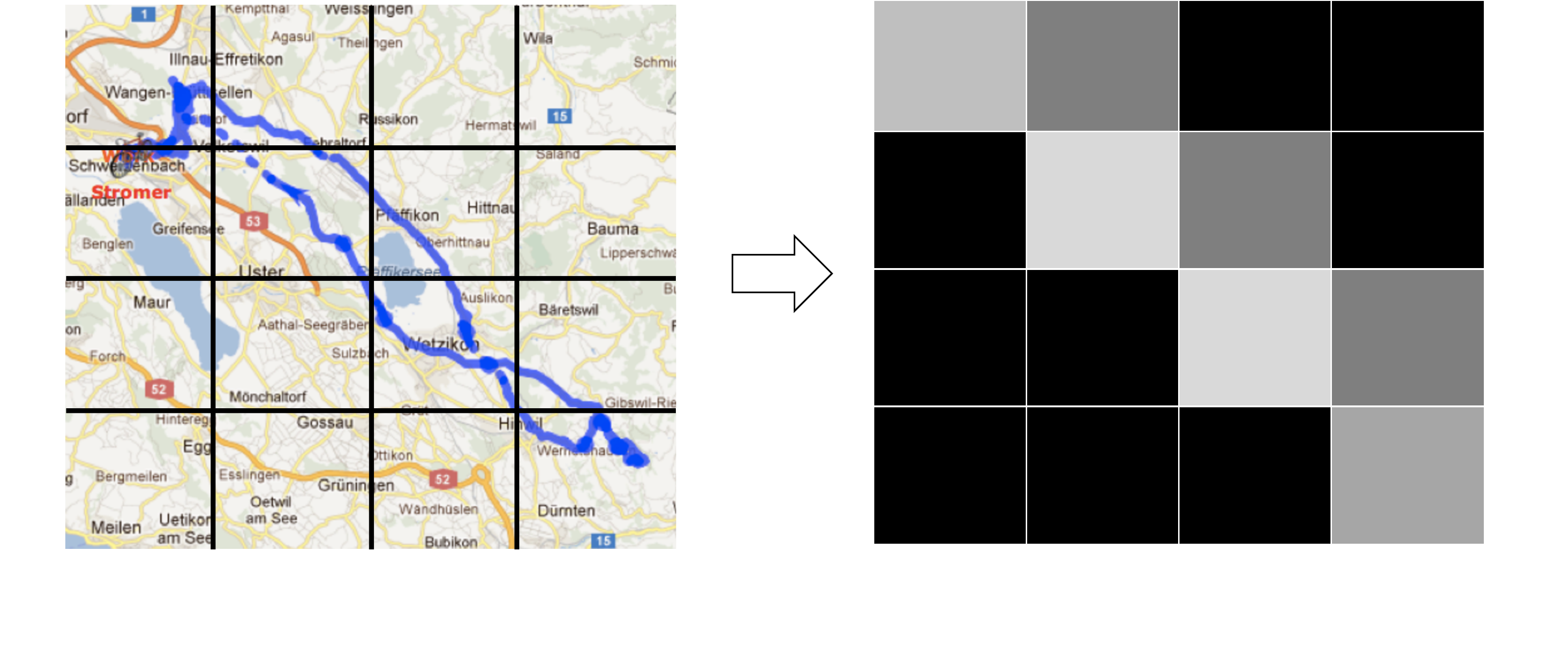}
        \vspace{-2em}
        \caption{Trajectory image example.}
        \label{fig:trajectory_image}
    \end{subfigure}
    \begin{subfigure}[t]{0.69\linewidth}
        \includegraphics[width=1\linewidth]{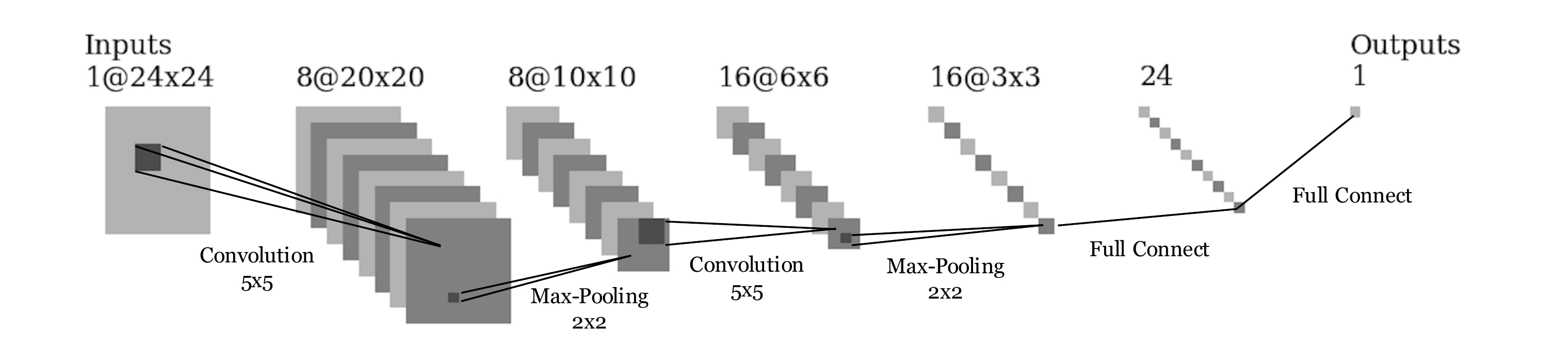}
        \vspace{-2em}
        \caption{Structure (City area is split to $24\times 24$. This figure is an example of day-level CNN and thus the input shape is $1@24\times24$; the input shape of car-level CNN is $7@24\times24$ if $7$-days traces are used.)}
        \label{fig:cnn_structure}
    \end{subfigure}
    \vspace{-1em}
    \caption{Convolutional neural network.}
    \vspace{-1em}
\end{figure*}



\textit{Trajectory Image.}  To map a car's traces into a gray-scale trajectory image, like the coverage measurements previously mentioned, we also split the whole city area into $M \times N$ grids, which can be seen as a $M \times N$ image. Then, for each grid, i.e. pixel, according to a car's traces, we count how much time that the car stays at the pixel and then set the lightness to be proportional to the stay time. We also set a threshold $T$: if the stay time of a pixel is larger than $T$, the color is set to white (255). The detailed calculation of a pixel's gray-scale color is as follows:
\begin{equation}
\setlength{\belowdisplayskip}{.5em} \setlength{\belowdisplayshortskip}{0pt}
\setlength{\abovedisplayskip}{.5em} \setlength{\abovedisplayshortskip}{0pt}
    \textstyle
    color_{i,j} = \min(\frac{t_{i,j}}{T}, 1) \cdot 255
\end{equation}
where $color_{i,j}$ is the color set for the pixel $\{i,j\}$, and $t_{i,j}$ is the stay time of the car at pixel $\{i,j\}$. We set $T$ to one hour in the experiment. Figure~\ref{fig:trajectory_image} shows an example of a car's raw trace and the mapped trajectory image.

\textit{Convolutional Neural Network.} To effectively extract the car driving patterns from the trajectory image, we build a neural network with several convolutional layers, which have been verified very efficient in various image-based classification tasks~\cite{lecun1998gradient,krizhevsky2012imagenet}. The network structure is shown in Figure~\ref{fig:cnn_structure}. More specifically, the CNN structure consists of the combinations of convolutional and max-pooling layers for two iterations, followed by one fully connect layer and finally the output layer with a sigmoid function to predict the ridesourcing probability. Our CNN structure can be seen as a simplified version of the deep CNN structure proposed by \cite{krizhevsky2012imagenet} for the ImageNet competition. We reduce the number of layers because fewer training instances exist in our case compared to ImageNet, and thus too many layers may increase the risk of overfitting. To further relieve the overfitting effect, we adopt the \textit{dropout} technique on the fully connect layer with a dropout probability of 0.5~\cite{srivastava2014dropout}. 

\textit{Ensemble of Day- and Car-level CNN.} 
In reality, we will usually use one car's (recent) several days' traces to decide whether it is used as ridesourcing or not. Hence, our CNN classifier needs to deal with the input of several days' trajectory images. Two aggregation methods are proposed to address this issue.

(1) \textit{Day-level CNN.} In the first method, we let the input of the CNN is one-day trajectory image. Suppose we have $K$-day traces for one car, then $K$ predictions exist. While some predictions may be conflicted, we adopt the widely-used average aggregation~\cite{zhou2012ensemble} to make the final decision, i.e., its probability of being a ridesourcing car is the mean probability of the $K$ days.

(2) \textit{Car-level CNN.} In the second method, we directly encode one car's $K$-day trajectory images into one $K$-channel image, i.e., the CNN input dimensions are $M \times N \times K$. Then, this CNN can directly output a car's probability of being ridesourcing or not.

Comparing the two aggregation methods, the advantage of the day-level CNN is that it has more training instances ($K$ times compared to car-level) and fewer trainable network parameters, and thus can better avoid overfitting. The advantage of the car-level CNN is its ability to automatically extract the features across different days' trajectories. We thus keep both of the classifiers in the co-training process, and construct the final CNN as an ensemble of day- and car-level CNN with average aggregation~\cite{zhou2012ensemble}.

\setlength{\textfloatsep}{5pt}
\begin{algorithm}[t]
\footnotesize
    \SetKwInOut{Input}{Input}
    \SetKwInOut{Output}{Output}
    \Input{$\mathcal C_r$: detected ridesourcing cars from knowledge transfer;\\
    $\mathcal C_n$: detected non-ridesourcing cars from knowledge transfer;\\
    $\mathcal C^*$: whole set of candidate cars;\\
    \textit{RF}: random forest classifier;\\
    \textit{CNN}: convolutional neural network classifier;\\
    $\delta$: confidence threshold for co-training.}
    \Output{Final classifier for ridesourcing car detection.}

    $\mathcal C' = C^* \setminus{(\mathcal C_r \cup \mathcal C_n)}$ \;
    \textit{update} $=$ \textit{true}\;
    \While{\textit{update}} {
    training \textit{RF} with $\{\mathcal C_r, \mathcal C_n\}$\;
    training \textit{CNN} with $\{\mathcal C_r, \mathcal C_n\}$\;
    $\mathcal C^r_r = $ ridesourcing cars detected by \textit{RF} from $\mathcal C'$ with conf. $>\delta$\;
    $\mathcal C^r_n = $ other cars detected by \textit{RF} from $\mathcal C'$ with conf. $>\delta$\;
    $\mathcal C^c_r = $ ridesourcing cars detected by \textit{CNN} from $\mathcal C'$ with conf. $>\delta$\;
    $\mathcal C^c_n = $ other cars detected by \textit{CNN} from $\mathcal C'$ with conf. $>\delta$\;
    $\mathcal C_r = \mathcal C_r \cup \mathcal C^r_r \cup \mathcal C^c_r$\;
    $\mathcal C_n = \mathcal C_n \cup \mathcal C^r_n \cup \mathcal C^c_n$\;
    $\mathcal C' = \mathcal C' \setminus{(\mathcal C_r \cup \mathcal C_n)}$\;
    \If{$\mathcal C^r_r \cup \mathcal C^r_n \cup \mathcal C^c_r \cup \mathcal C^c_n == \phi$} {
    \textit{update} $=$ \textit{false}\;
    }
    }
    \Return an ensemble classifier of \textit{RF} and \textit{CNN} by averaging their classification confidence\;
    \caption{Co-training Process}
    \label{alg:co_training}
\end{algorithm}

\vspace{+.5em}
\textbf{Co-training Process}

Algorithm~\ref{alg:co_training} shows the pseudocode of the co-training process used in our framework. 
Briefly, the algorithm begins with the input of the highly confident ridesourcing/other cars detected by Stage~1, and then iteratively refine both RF and CNN by adding newly confident ridesourcing/other cars labeled from the previous iteration. More specifically, in each iteration, the cars with confidence $>\delta$ are added into the training set. The algorithm terminates when no new cars can be added into the training set. We then use the ensemble of RF and CNN by averaging their classification confidence to determine the final label of a candidate car.





%% file: evaluation.tex
\section{Evaluation} 
\label{sec:evaluation}

In this section, we evaluate our proposed method on real car, taxi and bus traces in Shanghai. 

\subsection{Datasets} 
\label{sub:datasets}

Three datasets used in the evaluation purpose are as follows:

\textbf{SH-CAR}: This dataset includes the GPS traces of about 10,000 cars in Shanghai from 2016/05/01 to 2016/05/11, within which some cars may be ridesourcing while others are not. As we focus on identifying routine ridesourcing cars, we use the car traces in workdays during this time span, i.e., 2016/05/03--2016/05/06 and 2016/05/09--2016/05/11, a total of 7 days.

\textbf{SH-TAXI}: This dataset includes the GPS traces of about 4,400 taxis in Shanghai for 7 days, 2007/02/01--2007/02/07, which belongs to the SUVnet-Trace project~\cite{shanghaiTaxi}. 

\textbf{SH-BUS}: This dataset includes the GPS traces of about 2,800 buses in Shanghai for 7 days, 2007/03/01--2007/03/07, also provided by the SUVnet-Trace project~\cite{shanghaiTaxi}.

As expected, the time spans of the SH-TAXI and SH-BUS cannot align with SH-CAR, which reflects the challenges of knowledge transfer in real-life scenarios.

\subsection{Experiment Design} 
\label{sub:experiment_design}

To conduct the experiments, in Stage 1, we use SH-TAXI and SH-BUS as training data to learn RF and identify high-confident ridesourcing/other cars in SH-CAR. Then, in Stage 2, we run the co-training module to refine two classifiers, RF and CNN. Finally, we determine whether a car is ridesourcing or not using the ensemble of the co-trained RF and CNN. More specifically, we split the Shanghai city area into a $24\times 24$ grid when calculating the coverage matrix and trajectory image. The max-pooling operation used in Algorithm~\ref{alg:coverage_diff} is set to map a $2\times 2$ sub-matrix; $\delta$ in Algorithm~\ref{alg:co_training} is set to 0.9. Detailed CNN structure is shown in Figure~\ref{fig:cnn_structure}, and the number of trees in RF is set to 100.

To evaluate the performance, we recruit five university students to label randomly selected 600 cars in SH-CAR as ridesourcing or not as test data. All the participants are familiar with Shanghai urban area. Each car is annotated by all of them, and the final label of each car, considered as `ground truth', is determined by majority voting ($\ge$ three participants)~\cite{yuen2011survey}, leading to 290/310 ridesourcing/other cars. To verify the generality of our method, we do not include these 600 cars in co-training.

For assessment, we use common classification evaluation metrics, \textit{AUC}\footnote{\url{en.wikipedia.org/wiki/Receiver_operating_characteristic\#Area_under_the_curve}} and \textit{accuracy}\footnote{\url{en.wikipedia.org/wiki/Accuracy_and_precision\#In_binary_classification}; the decision boundary for measuring accuracy is always set to 0.5 for our method and all baselines.}. In addition, we also use the metric of \textit{top-$k\%$ precision} (abbr. \textit{t $k\%$-prec.}), which measures of the ratio of true ridesourcing cars among the top $k\%$ of cars with the highest confidence. This metric is important in many real-life settings. For example, if we want to identify suspicious unauthorized ridesourcing cars for further investigations, the number of cars selected for inspection may be limited due to budget constraints. Then, the detection precision among the top confident ridesourcing cars is a key metric.

Our experiment platform is an ordinary laptop with i7-6700HQ (2.60 GHz), 8GB RAM, and Nvidia GTX960M (4GB VRAM). We use Python~2.7 with scikit-learn\footnote{\url{scikit-learn.org}} and tensorflow\footnote{\url{www.tensorflow.org}} on Ubuntu 14.04 to implement our methods and baselines. 

\subsection{Baselines} 
\label{sub:baselines}

We include three types of baselines for comparison in our experiments. \textit{Src-Supervised} baselines train the classier in the source domain SH-TAXI\&SH-BUS, and directly apply the classifier to detect ridesourcing cars in the target domain SH-CAR.
\begin{itemize}
	\item \textit{Src-RF} trains RF (Sec.~\ref{sub:taxi_knowledge_transfer}) on SH-TAXI and SH-BUS. 
	\item \textit{Src-CNN} trains CNN (Sec.~\ref{sub:co_training}) on SH-TAXI and SH-BUS.
\end{itemize}

\textit{Tgt-Supervised} learning baselines directly use the labels in the target domain (600 labeled cars in SH-CAR) for training, without leveraging the source domain data. This kind of methods is much more difficult to be employed in real life than our approach due to the difficulty in obtaining actual ridesourcing labels. The results are obtained through 5-fold cross validation on the 600 labeled cars.
\begin{itemize}
	\item \textit{Tgt-RF} trains RF based on labeled SH-CAR.
	\item \textit{Tgt-CNN} trains CNN based on labeled SH-CAR.
	\item \textit{Tgt-RF\&CNN} uses the ensemble of RF and CNN trained on labeled SH-CAR for ridesourcing detection.
\end{itemize}

\textit{Transfer} learning baselines use the knowledge in the source domain SH-TAXI\&SH-BUS to do the ridesourcing detection task in the target domain SH-CAR.
\begin{itemize}
	\item \textit{TrAdaBoost}~\cite{dai2007boosting} is an instance-based transfer learning algorithm which combines the labeled instances in both target and source domains for training, with a mechanism to properly assigning instance training weights. As TrAdaBoost needs a small number of labels in the target domain as input, we suppose that $10\%$ of the labeled SH-CAR data (60 cars) are known to TrAdaBoost. The basic learner is RF.
	\item \textit{Tr-RF-self} has the same Stage~1 as our method, while in Stage~2, it uses only RF for self-training~\cite{chapelle2009semi}.
	\item \textit{Tr-CNN-self} has the same Stage~1 as our method, while in Stage~2, it uses only CNN for self-training. 
\end{itemize}




\subsection{Experiment Results} 
\label{sub:experiment_results}

We present our experiment results in three parts. First, we describe the main evaluation outcomes by comparing our method to various baselines. Second, we look deeper into the shared features we extracted in Sec.~\ref{sub:taxi_knowledge_transfer}, verifying the effectiveness of each shared feature and how the noises in taxi data, if any, will affect the shared features. Finally, we analyze the running time performance.

\vspace{+.5em}
\textbf{Overall Results}

The experiment results of our method and the baselines are shown in Table~\ref{tbl:overall_evaluation_results}. First, our method outperforms the other three transfer learning baselines. Among them, \textit{Tr-RF-self} and \textit{Tr-CNN-self} are two simplified variants of our method, and thus our improvement is expected. Our method introduces two different types of classifiers in Stage~2 and uses a co-training process, while the two baselines only use one classifier with self-training. Regarding \textit{TrAdaBoost}, even given 10\% of labels of the target domain data, it still performs worse than our method. The probable reason is that the assumption that the tasks in both domains are identical does not hold in our problem, as our tasks in two domains, although related to each other, are not exactly the same. 

\begin{table}[t]
\centering
\scriptsize
\begin{tabular}{lccccc}
\hline
\textbf{}             & \textbf{AUC} & \textbf{Accuracy} & \textbf{t5\%-prec.} & \textbf{t10\%-prec.} & \textbf{Tgt-Label?} \\ \hline
\textbf{Manual label} &   ---        & \parbox{1.3cm}{\centering 0.844 (mean)\\0.845 (median)}          &      ---             &   ---     & ---       \\ \hline
\multicolumn{5}{l}{\textbf{Src-Supervised}}                                                      \\
\textit{Src-RF}        &  0.789       &   0.427           &  0.911               &  0.850        &  no\\
\textit{Src-CNN}       &  0.613       &   0.601           &  0.700               &  0.817        &  no\\ \hline
\multicolumn{5}{l}{\textbf{Tgt-Supervised}}                                                       \\
\textit{Tgt-RF}         &  0.900       &   0.826           &  0.933               &  0.894       & yes\\
\textit{Tgt-CNN}        &  0.893       &   0.823           &  0.936               &  0.852       & yes\\
\textit{Tgt-RF\&CNN}    &  0.920       &   0.855           &  0.971               &  0.903       & yes\\ \hline
\multicolumn{5}{l}{\textbf{Transfer}}                                           \\
\textit{TrAdaBoost}    &  0.873       &   0.804           &  0.927               &  0.895     & some\\
\textit{Tr-RF-self}        &  0.822        &   0.786           &  0.900              &  0.867    & no\\
\textit{Tr-CNN-self}       &  0.893        &   0.790           &  0.800              &  0.833    & no\\
\textit{Our method}       &  0.910        &   0.852           &  0.967              &  0.900     & no\\ \hline
\end{tabular}
\caption{Overall evaluation results. (`\textit{Tgt-Label}': whether the method needs labeled data in the target domain)}
\label{tbl:overall_evaluation_results}
\vspace{-1.5em}
\end{table}

Comparing to \textit{Tgt-Supervised} baselines, our method achieves a comparable accuracy to the ensemble classifier \textit{Tgt-RF\&CNN}, while outperforming the single supervised classifier, \textit{Tgt-RF} and \textit{Tgt-CNN}. This is a really exciting result, as our method can achieve a similar performance as the supervised classifiers  without using any labeled data in the target domain. Considering the difficulty of obtaining ridesourcing car labels in reality, our method based on knowledge transferred from public transportation has a notable advantage over the supervised learning approaches in practice.

All the  \textit{Src-Supervised} baselines perform rather poorly in AUC and accuracy.\footnote{The accuracy of \textit{Src-RF} is even worse than random guess (accuracy is $0.5$),  simply because the classification decision boundary of 0.5 is inappropriate for \textit{Src-RF}. AUC is a more useful metric to check the overall prediction ability of \textit{Src-RF}. } This indicates that, although taxis share similar patterns with some ridesourcing cars, still a large number of ridesourcing cars cannot be directly detected by the classifier trained on the taxi data. This stresses the necessity of introducing Stage~2 in our method. However, \textit{Src-RF} performs fairly well in terms of \textit{top k\% precision}. Actually, the good performance of \textit{Src-RF} on \textit{top $k\%$ precision} is the basis for the validity of our method, because \textit{Src-RF} is exactly the classifier that we use to detect highly confident ridesourcing/other cars in Stage~1. We can also observe that for \textit{Src-CNN} (CNN trained on the source domain), \textit{top $k\%$ precision} is still poor, indicating that the features used by CNN (i.e., stay-time in each city cell) do not transfer well across source and target domains, perhaps due to the time misalignment issue. These comparisons show the importance of choosing proper shared features in Stage~1, and also verify the effectiveness of our selected shared features in Sec.~\ref{sub:taxi_knowledge_transfer}.

Finally, we compute the manual label accuracy, i.e., the percentage of a participant's labels that are same as majority voting. The manual label accuracy ranges from 0.820 to 0.858 for different participants, with the mean value of 0.844 and the median value of 0.845. Considering that our method can achieve a detection accuracy of 0.852, it actually performs as good as a human in the task of labeling ridesourcing cars.

\vspace{+.5em}
\textbf{Effectiveness of Shared Features}

\begin{figure}[t]
    \centering
    \begin{subfigure}[t]{0.48\linewidth}
        \includegraphics[width=1\linewidth]{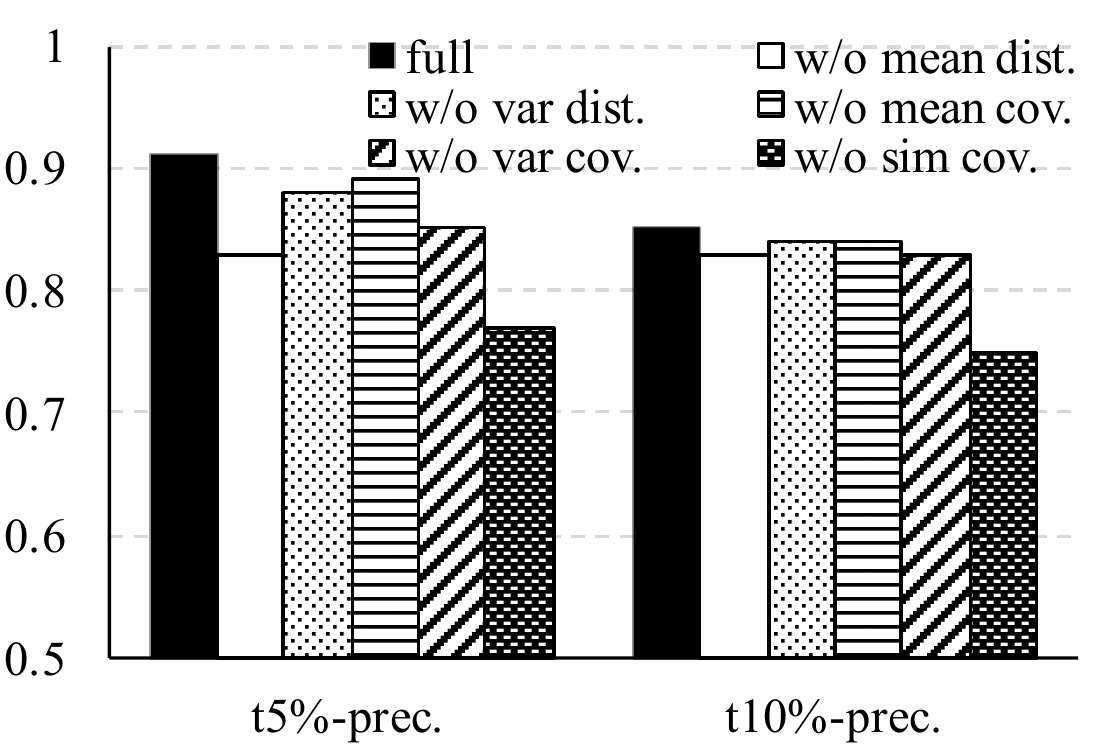}
        \caption{Leave-one-feature-out}
        \label{fig:eval_shared_features_leave_one_out}
    \end{subfigure}
    \begin{subfigure}[t]{0.48\linewidth}
        \includegraphics[width=1\linewidth]{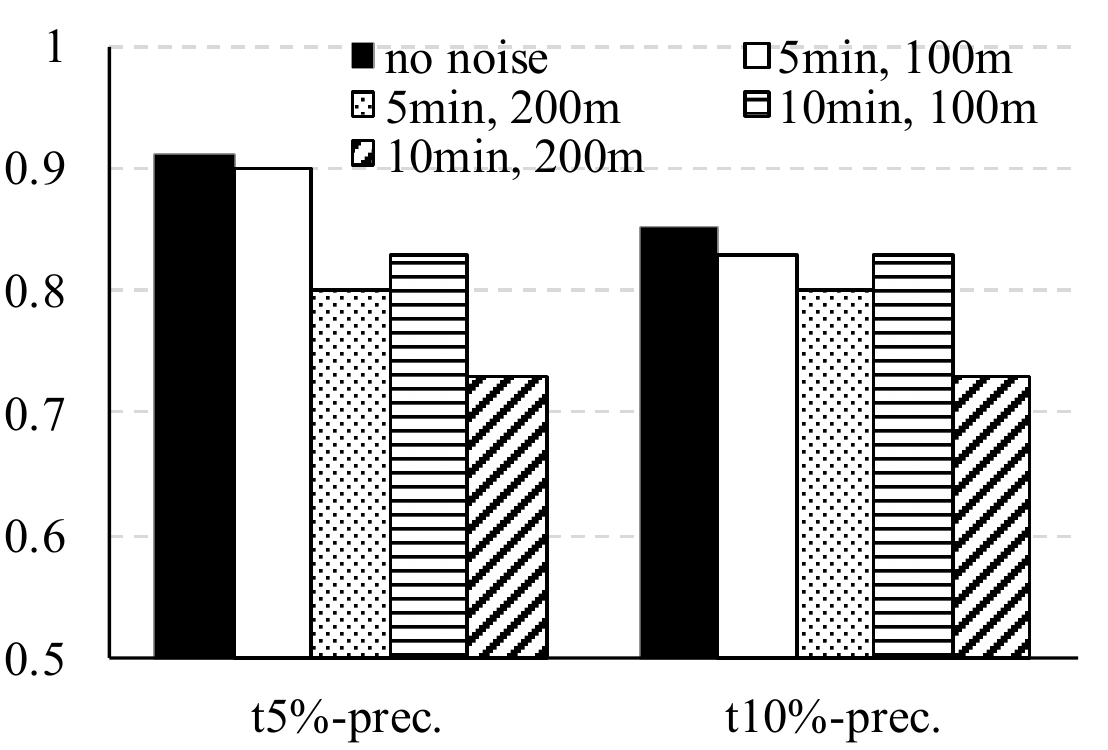}
        \caption{Spatio-temporal noises}
        \label{fig:eval_shared_features_noise}
    \end{subfigure}
    \vspace{-1em}
    \caption{Evaluation results on shared features in Stage 1.}
\end{figure}

Selecting appropriate shared features in Stage~1 is critical to the performance of our learning framework. Here, we conduct further analysis on the shared features identified in Sec.~\ref{sub:taxi_knowledge_transfer}. As the objective of using shared features in Stage~1 is to produce the initial set of highly confident ridesourcing/other labels, we focus on \textit{top $k\%$ precision}.  

First, we verify the effectiveness of each feature using the \textit{leave-one-feature-out} evaluation method~\cite{han2015alike}. In each iteration, we eliminate one feature from the input of the RF classifier, re-learn the RF on the source domain (SH-TAXI\&SH-BUS), and test on the target domain (600 labeled cars in SH-CAR). There are five feature types: \textit{mean/variance of distance} and \textit{mean/variance/similarity of coverage}. The results are shown in Figure~\ref{fig:eval_shared_features_leave_one_out}. We can see that the \textit{top $k\%$ precision} after removing any type of features is lower than the full model, which verifies the effectiveness of each type of the shared features. Particularly, we find that the coverage similarity contributes the most among all shared features. We also try replacing our robust similarity metric with cell-to-cell Jaccard similarity metric to measure the intra- and inter-day coverage similarity, and then the \textit{top $5\%$ precision} drops significantly from 0.91 to 0.75, further confirming that our proposed robust similarity metric plays a key role in shared features.

Recently, some open taxi data have added spatio-temporal perturbations due to privacy concerns~\cite{ChicagoTaxi}. Intuitively, such perturbations may degrade the effectiveness of our extracted shared features in identifying the high-confident ridesourcing cars in Stage~1. To assess the potential impact, we do a preliminary study to see how our shared features would be affected by the spatio-temporal noises. In particular, we reduce the temporal sampling rate as well as the spatial resolution in SH-TAXI, following the idea in~\cite{ChicagoTaxi}. This can introduce a noise of `\textit{$X$ min, $Y$ m}', meaning that the taxi trace sampling rate is one location point every $X$ minutes, and the location point is randomly placed within an area of a Y-meter radius from the actual location. As shown in Figure~\ref{fig:eval_shared_features_noise}, a small amount of noise, e.g., `\textit{5min, 100m}' would not affect the effectiveness of the shared features significantly. However, as the noise increases, the performance of our shared features learned from the noisy taxi data degrades. In the taxi open data of Chicago~\cite{ChicagoTaxi}, the temporal sampling rate is 15 minutes and the spatial resolution is \textit{census tract} (an area whose length and width are often longer than 500m). To deal with such a level of noise, future work on designing more appropriate shared features is necessary.

\vspace{+.5em}
\textbf{Running Time Performance}

In our method, most of the running time is spent on CNN training in Stage~2, which may take up to 10 minutes in one iteration. As a comparison, RF training only takes about one second for each iteration. In our experiment, the co-training process usually terminates at the 5th or 6th iteration, and thus the totally time consumption of our method is about one hour. Since training the classifier to detect ridesourcing cars can be conducted offline, such time consumption is feasible for real-life deployment.



%% file: related_work.tex
\vspace{-.5em}
\section{Related Work} 
\label{sec:related_work}

Trajectory mining is a hot research topic nowadays, where a spectrum of applications have been successfully developed, such as city-scale map creation~\cite{chencity}, human transportation mode detection~\cite{zheng2010understanding} and crowd mobility prediction~\cite{wang2015regularity}. A nice survey can be found in~\cite{zheng2015trajectory}. More specifically, as the taxi traces are perhaps the most easily accessible large-scale open data for trajectory mining~\cite{beijingTaxi,shanghaiTaxi,NYTaxi,romaTaxi,SFTaxi}, plenty of research studies are conducted by using it as an important data source, e.g., anomaly detection~\cite{zhang2011ibat}, environment monitoring~\cite{zheng2013u}, bus route planning~\cite{chen2014b}, travel time estimation~\cite{wang2014travel} and personalized trip navigation~\cite{chen2015tripplanner}; a comprehensive survey on taxi trajectory mining can be found in~\cite{castro2013taxi}. 

Similar to our research topic, i.e., vehicle classification, previous studies have used GPS trajectories to classify vehicles into delivery trucks or passenger cars~\cite{sun2013vehicle}. In addition to different research objectives between ours and \cite{sun2013vehicle}, the method used in \cite{sun2013vehicle} is supervised learning, while our method belongs to transfer learning~\cite{pan2010survey}, which can address the difficulty in obtaining the labeled ridesourcing dataset in reality. Note that in traditional transportation research literature, besides GPS sensors, a variety of other sensors (e.g., radar, acoustic and computer vision-based sensors) are also used to vehicle classification; however, such sensors are generally deployed in fixed locations and expensive to be applied in a large scale~\cite{avery2004length}. 

More recently, with the rapid development of ridesourcing market, both industry and academic researchers have started devoting efforts to mining ridesourcing data~\cite{sharedMobility}. Since 2015, Didi has published two yearly (2015 and 2016) reports on China smart transportation based on the trip data in its platform~\cite{didi_big_data_report}. Chen et al.~\cite{chen2017understanding} leverage supervised ensemble learning to identify ridesplitting behavior (e.g., hitch) from a set of ridesourcing trips considering the trip time, costs, waiting time, etc. Rayle et al.~\cite{rayle2016just} conduct a survey-based comparison of taxi and crowdsourcing services in San Francisco, summarizing both similarities and differences between the two services. While the research studies about ridesourcing are arising, to the best of our knowledge, we are the first one to design a method which can easily and widely detect ridesourcing cars from a large set of candidate cars based on their trajectories.

Our solution to detecting ridesourcing cars belongs to transfer learning~\cite{pan2010survey}, while also inspired by semi-supervised learning techniques~\cite{chapelle2009semi}, and more specifically, co-training~\cite{blum1998combining,goldman2000enhancing}. In our method, both random forest~\cite{liaw2002classification} and convolutional neural network~\cite{zhang2016deep,krizhevsky2012imagenet} classifiers are constructed for ridesourcing car identification, and the co-training techniques are employed to refine the classifiers iteratively to achieve better performance. The final classifier is an ensemble of the two classifiers~\cite{zhou2012ensemble}. Compared to traditional co-training processes that need a (small) number of labels, by transferring knowledge from public transportation open data, we do not need any ridesourcing labels for detection. 

%% file: conclusion.tex
\vspace{-1em}
\section{Conclusion} 
\label{sec:conclusion}

In this paper, we propose a method to detect ridesourcing cars from a pool of vehicles based on their trajectories, which may help to regulate the black market activities in the rapidly-growing ridesourcing industry. Since the licensed ridesourcing car trace data are generally unavailable to the public, and some cities may not even have legal ridesourcing services, we propose to transfer the knowledge from taxi and bus open data to ridesourcing car detection. Our experiment results show that, with no need of any pre-labeled ridesourcing car dataset, our method can achieve a comparable detection accuracy as the supervised learning methods.

In the future, we plan to try other popular deep neural network structures, like residual network~\cite{zhang2016deep}, for ridesourcing car detection. Besides, as some taxi data only include pick-ups and drop-offs~\cite{NYTaxi}, we would like to study the performance of our method if we need to first re-construct the taxis' trajectories from pick-ups and drop-offs. A more fine-grained ridesourcing activity classification, such as differentiating full-time and part-time ridesourcing, will also be explored. As discussed in the evaluation, addressing noisy taxi data is another research direction. Finally, we plan to investigate the theoretical properties, e.g., convergence, of our proposed transfer learning framework.


%% file: car_usage_classification_kdd.bbl

\begin{thebibliography}{00}


\ifx \showCODEN    \undefined \def \showCODEN     #1{\unskip}     \fi
\ifx \showDOI      \undefined \def \showDOI       #1{{\tt DOI:}\penalty0{#1}\ }
  \fi
\ifx \showISBNx    \undefined \def \showISBNx     #1{\unskip}     \fi
\ifx \showISBNxiii \undefined \def \showISBNxiii  #1{\unskip}     \fi
\ifx \showISSN     \undefined \def \showISSN      #1{\unskip}     \fi
\ifx \showLCCN     \undefined \def \showLCCN      #1{\unskip}     \fi
\ifx \shownote     \undefined \def \shownote      #1{#1}          \fi
\ifx \showarticletitle \undefined \def \showarticletitle #1{#1}   \fi
\ifx \showURL      \undefined \def \showURL       #1{#1}          \fi
\providecommand\bibfield[2]{#2}
\providecommand\bibinfo[2]{#2}
\providecommand\natexlab[1]{#1}
\providecommand\showeprint[2][]{arXiv:#2}

\bibitem[\protect\citeauthoryear{??}{DId}{2016}]%
        {DIdiOrderHacked}
 \bibinfo{year}{2016}\natexlab{}.
\newblock \bibinfo{title}{{Your Didi order may be hacked (in Chinese)}}.
\newblock
  \bibinfo{howpublished}{\url{http://www.bjd.com.cn/sd/hcr/201604/26/t20160426_11019444.html}}.
    (\bibinfo{year}{2016}).
\newblock
\newblock
\shownote{Accessed: 2017-01-15.}


\bibitem[\protect\citeauthoryear{??}{Did}{2017}]%
        {Didi}
 \bibinfo{year}{2017}\natexlab{}.
\newblock \bibinfo{title}{{Didi}}.
\newblock \bibinfo{howpublished}{\url{http://www.xiaojukeji.com/}}.
  (\bibinfo{year}{2017}).
\newblock
\newblock
\shownote{Accessed: 2017-01-15.}


\bibitem[\protect\citeauthoryear{??}{Iri}{2017}]%
        {Irish_bus_data}
 \bibinfo{year}{2017}\natexlab{}.
\newblock \bibinfo{title}{{Irish Bus Data.}}
\newblock
  \bibinfo{howpublished}{\url{http://www.transportforireland.ie/transitData/PT_Data.html}}.
    (\bibinfo{year}{2017}).
\newblock
\newblock
\shownote{Accessed: 2017-02-12.}


\bibitem[\protect\citeauthoryear{??}{Lyf}{2017}]%
        {Lyft}
 \bibinfo{year}{2017}\natexlab{}.
\newblock \bibinfo{title}{{Lyft}}.
\newblock \bibinfo{howpublished}{\url{https://www.lyft.com/}}.
  (\bibinfo{year}{2017}).
\newblock
\newblock
\shownote{Accessed: 2017-01-15.}


\bibitem[\protect\citeauthoryear{??}{Dan}{2017}]%
        {Dangerous_gypsy_cab}
 \bibinfo{year}{2017}\natexlab{}.
\newblock \bibinfo{title}{{One of the Most Dangerous Jobs in New York: Gypsy
  Cab Driver.}}
\newblock \bibinfo{howpublished}{\url{http://www.taxi-library.org/marosi.htm}}.
    (\bibinfo{year}{2017}).
\newblock
\newblock
\shownote{Accessed: 2017-02-12.}


\bibitem[\protect\citeauthoryear{??}{Ube}{2017}]%
        {Uber}
 \bibinfo{year}{2017}\natexlab{}.
\newblock \bibinfo{title}{{Uber}}.
\newblock \bibinfo{howpublished}{\url{https://www.uber.com}}.
  (\bibinfo{year}{2017}).
\newblock
\newblock
\shownote{Accessed: 2017-01-15.}


\bibitem[\protect\citeauthoryear{Avery, Wang, and Rutherford}{Avery
  et~al\mbox{.}}{2004}]%
        {avery2004length}
\bibfield{author}{\bibinfo{person}{Ryan~P Avery}, \bibinfo{person}{Yinhai
  Wang}, {and} \bibinfo{person}{G~Scott Rutherford}.}
  \bibinfo{year}{2004}\natexlab{}.
\newblock \showarticletitle{Length-based vehicle classification using images
  from uncalibrated video cameras}. In \bibinfo{booktitle}{{\em Proceedings of
  ITSC}}. IEEE, \bibinfo{pages}{737--742}.
\newblock


\bibitem[\protect\citeauthoryear{Blum and Mitchell}{Blum and Mitchell}{1998}]%
        {blum1998combining}
\bibfield{author}{\bibinfo{person}{Avrim Blum} {and} \bibinfo{person}{Tom
  Mitchell}.} \bibinfo{year}{1998}\natexlab{}.
\newblock \showarticletitle{Combining labeled and unlabeled data with
  co-training}. In \bibinfo{booktitle}{{\em Proceedings of the eleventh annual
  conference on Computational learning theory}}. ACM, \bibinfo{pages}{92--100}.
\newblock


\bibitem[\protect\citeauthoryear{Bracciale, Bonola, Loreti, Bianchi, Amici, and
  Rabuffi}{Bracciale et~al\mbox{.}}{2014}]%
        {romaTaxi}
\bibfield{author}{\bibinfo{person}{Lorenzo Bracciale}, \bibinfo{person}{Marco
  Bonola}, \bibinfo{person}{Pierpaolo Loreti}, \bibinfo{person}{Giuseppe
  Bianchi}, \bibinfo{person}{Raul Amici}, {and} \bibinfo{person}{Antonello
  Rabuffi}.} \bibinfo{year}{2014}\natexlab{}.
\newblock \bibinfo{title}{{CRAWDAD} dataset roma/taxi (v. 2014-07-17)}.
\newblock \bibinfo{howpublished}{Downloaded from
  \url{http://crawdad.org/roma/taxi/20140717}}.   (\bibinfo{date}{July}
  \bibinfo{year}{2014}).
\newblock
\showDOI{%
\url{http://dx.doi.org/10.15783/C7QC7M}}


\bibitem[\protect\citeauthoryear{Castro, Zhang, Chen, Li, and Pan}{Castro
  et~al\mbox{.}}{2013}]%
        {castro2013taxi}
\bibfield{author}{\bibinfo{person}{Pablo~Samuel Castro},
  \bibinfo{person}{Daqing Zhang}, \bibinfo{person}{Chao Chen},
  \bibinfo{person}{Shijian Li}, {and} \bibinfo{person}{Gang Pan}.}
  \bibinfo{year}{2013}\natexlab{}.
\newblock \showarticletitle{From taxi GPS traces to social and community
  dynamics: A survey}.
\newblock \bibinfo{journal}{{\em ACM Computing Surveys (CSUR)\/}}
  \bibinfo{volume}{46}, \bibinfo{number}{2} (\bibinfo{year}{2013}),
  \bibinfo{pages}{17}.
\newblock


\bibitem[\protect\citeauthoryear{Chapelle, Scholkopf, and Zien}{Chapelle
  et~al\mbox{.}}{2009}]%
        {chapelle2009semi}
\bibfield{author}{\bibinfo{person}{Olivier Chapelle}, \bibinfo{person}{Bernhard
  Scholkopf}, {and} \bibinfo{person}{Alexander Zien}.}
  \bibinfo{year}{2009}\natexlab{}.
\newblock \showarticletitle{Semi-Supervised Learning (Chapelle, O. et al.,
  Eds.; 2006)[Book reviews]}.
\newblock \bibinfo{journal}{{\em IEEE Transactions on Neural Networks\/}}
  \bibinfo{volume}{20}, \bibinfo{number}{3} (\bibinfo{year}{2009}),
  \bibinfo{pages}{542--542}.
\newblock


\bibitem[\protect\citeauthoryear{Chen, Lu, Huang, Yang, Gunopulos, and
  Guibas}{Chen et~al\mbox{.}}{2016}]%
        {chencity}
\bibfield{author}{\bibinfo{person}{Chen Chen}, \bibinfo{person}{Cewu Lu},
  \bibinfo{person}{Qixing Huang}, \bibinfo{person}{Qiang Yang},
  \bibinfo{person}{Dimitrios Gunopulos}, {and} \bibinfo{person}{Leonidas
  Guibas}.} \bibinfo{year}{2016}\natexlab{}.
\newblock \showarticletitle{City-Scale Map Creation and Updating using GPS
  Collections}. In \bibinfo{booktitle}{{\em Proceedings of KDD}}. ACM.
\newblock


\bibitem[\protect\citeauthoryear{Chen, Zhang, Guo, Ma, Pan, and Wu}{Chen
  et~al\mbox{.}}{2015}]%
        {chen2015tripplanner}
\bibfield{author}{\bibinfo{person}{Chao Chen}, \bibinfo{person}{Daqing Zhang},
  \bibinfo{person}{Bin Guo}, \bibinfo{person}{Xiaojuan Ma},
  \bibinfo{person}{Gang Pan}, {and} \bibinfo{person}{Zhaohui Wu}.}
  \bibinfo{year}{2015}\natexlab{}.
\newblock \showarticletitle{TripPlanner: Personalized trip planning leveraging
  heterogeneous crowdsourced digital footprints}.
\newblock \bibinfo{journal}{{\em IEEE Transactions on Intelligent
  Transportation Systems\/}} \bibinfo{volume}{16}, \bibinfo{number}{3}
  (\bibinfo{year}{2015}), \bibinfo{pages}{1259--1273}.
\newblock


\bibitem[\protect\citeauthoryear{Chen, Zhang, Li, and Zhou}{Chen
  et~al\mbox{.}}{2014}]%
        {chen2014b}
\bibfield{author}{\bibinfo{person}{Chao Chen}, \bibinfo{person}{Daqing Zhang},
  \bibinfo{person}{Nan Li}, {and} \bibinfo{person}{Zhi-Hua Zhou}.}
  \bibinfo{year}{2014}\natexlab{}.
\newblock \showarticletitle{B-Planner: Planning bidirectional night bus routes
  using large-scale taxi GPS traces}.
\newblock \bibinfo{journal}{{\em IEEE Transactions on Intelligent
  Transportation Systems\/}} \bibinfo{volume}{15}, \bibinfo{number}{4}
  (\bibinfo{year}{2014}), \bibinfo{pages}{1451--1465}.
\newblock


\bibitem[\protect\citeauthoryear{Chen, Zahiri, and Zhang}{Chen
  et~al\mbox{.}}{2017}]%
        {chen2017understanding}
\bibfield{author}{\bibinfo{person}{Xiqun~Michael Chen}, \bibinfo{person}{Majid
  Zahiri}, {and} \bibinfo{person}{Shuaichao Zhang}.}
  \bibinfo{year}{2017}\natexlab{}.
\newblock \showarticletitle{Understanding ridesplitting behavior of on-demand
  ride services: An ensemble learning approach}.
\newblock \bibinfo{journal}{{\em Transportation Research Part C: Emerging
  Technologies\/}}  \bibinfo{volume}{76} (\bibinfo{year}{2017}),
  \bibinfo{pages}{51--70}.
\newblock


\bibitem[\protect\citeauthoryear{Dai, Yang, Xue, and Yu}{Dai
  et~al\mbox{.}}{2007}]%
        {dai2007boosting}
\bibfield{author}{\bibinfo{person}{Wenyuan Dai}, \bibinfo{person}{Qiang Yang},
  \bibinfo{person}{Gui-Rong Xue}, {and} \bibinfo{person}{Yong Yu}.}
  \bibinfo{year}{2007}\natexlab{}.
\newblock \showarticletitle{Boosting for transfer learning}. In
  \bibinfo{booktitle}{{\em Proceedings of ICML}}. ACM,
  \bibinfo{pages}{193--200}.
\newblock


\bibitem[\protect\citeauthoryear{Daily}{Daily}{2016}]%
        {DidiKill}
\bibfield{author}{\bibinfo{person}{Southern~Metropolis Daily}.}
  \bibinfo{year}{2016}\natexlab{}.
\newblock \bibinfo{title}{{Shenzhen female teacher is killed by Didi driver (in
  Chinese)}}.
\newblock
  \bibinfo{howpublished}{\url{http://epaper.oeeee.com/epaper/A/html/2016-05/04/content_33327.htm}}.
    (\bibinfo{year}{2016}).
\newblock
\newblock
\shownote{Accessed: 2017-01-15.}


\bibitem[\protect\citeauthoryear{Davidson}{Davidson}{2016}]%
        {austin_black_market}
\bibfield{author}{\bibinfo{person}{John~Daniel Davidson}.}
  \bibinfo{year}{2016}\natexlab{}.
\newblock \bibinfo{title}{{Black Market Ride-Sharing Explodes In Austin After
  Voters Drive Out Uber And Lyft.}}
\newblock
  \bibinfo{howpublished}{\url{http://thefederalist.com/2016/05/23/black-market-ride-sharing-uber-lyft/}}.
    (\bibinfo{year}{2016}).
\newblock
\newblock
\shownote{Accessed: 2017-02-15.}


\bibitem[\protect\citeauthoryear{Didi and CBNData}{Didi and CBNData}{2016}]%
        {didi_big_data_report}
\bibfield{author}{\bibinfo{person}{Didi} {and} \bibinfo{person}{CBNData}.}
  \bibinfo{year}{2016}\natexlab{}.
\newblock \bibinfo{title}{{Smart Transportation Big Data Report (in Chinese).}}
\newblock \bibinfo{howpublished}{\url{http://www.imxdata.com/archives/20017}}.
   (\bibinfo{year}{2016}).
\newblock
\newblock
\shownote{Accessed: 2017-02-17.}


\bibitem[\protect\citeauthoryear{Digital}{Digital}{2017}]%
        {ChicagoTaxi}
\bibfield{author}{\bibinfo{person}{Chicago Digital}.}
  \bibinfo{year}{2017}\natexlab{}.
\newblock \bibinfo{title}{Chicago Taxi Data}.
\newblock
  \bibinfo{howpublished}{\url{http://digital.cityofchicago.org/index.php/chicago-taxi-data-released/
  }}.   (\bibinfo{year}{2017}).
\newblock
\newblock
\shownote{Accessed: 2017-01-15.}


\bibitem[\protect\citeauthoryear{Goldman and Zhou}{Goldman and Zhou}{2000}]%
        {goldman2000enhancing}
\bibfield{author}{\bibinfo{person}{Sally Goldman} {and} \bibinfo{person}{Yan
  Zhou}.} \bibinfo{year}{2000}\natexlab{}.
\newblock \showarticletitle{Enhancing supervised learning with unlabeled data}.
  In \bibinfo{booktitle}{{\em ICML}}. \bibinfo{pages}{327--334}.
\newblock


\bibitem[\protect\citeauthoryear{Han, Wang, Crespi, Park, and Cuevas}{Han
  et~al\mbox{.}}{2015}]%
        {han2015alike}
\bibfield{author}{\bibinfo{person}{Xiao Han}, \bibinfo{person}{Leye Wang},
  \bibinfo{person}{Noel Crespi}, \bibinfo{person}{Soochang Park}, {and}
  \bibinfo{person}{{\'A}ngel Cuevas}.} \bibinfo{year}{2015}\natexlab{}.
\newblock \showarticletitle{Alike people, alike interests? Inferring interest
  similarity in online social networks}.
\newblock \bibinfo{journal}{{\em Decision Support Systems\/}}
  \bibinfo{volume}{69} (\bibinfo{year}{2015}), \bibinfo{pages}{92--106}.
\newblock


\bibitem[\protect\citeauthoryear{Hunter, Abbeel, and Bayen}{Hunter
  et~al\mbox{.}}{2014}]%
        {hunter2014path}
\bibfield{author}{\bibinfo{person}{Timothy Hunter}, \bibinfo{person}{Pieter
  Abbeel}, {and} \bibinfo{person}{Alexandre Bayen}.}
  \bibinfo{year}{2014}\natexlab{}.
\newblock \showarticletitle{The path inference filter: model-based low-latency
  map matching of probe vehicle data}.
\newblock \bibinfo{journal}{{\em IEEE Transactions on Intelligent
  Transportation Systems\/}} \bibinfo{volume}{15}, \bibinfo{number}{2}
  (\bibinfo{year}{2014}), \bibinfo{pages}{507--529}.
\newblock


\bibitem[\protect\citeauthoryear{Jetcheva, Hu, PalChaudhuri, Saha, and
  Johnson}{Jetcheva et~al\mbox{.}}{2003}]%
        {rice-ad_hoc_city-20030911}
\bibfield{author}{\bibinfo{person}{Jorjeta~G. Jetcheva},
  \bibinfo{person}{Yih-Chun Hu}, \bibinfo{person}{Santashil PalChaudhuri},
  \bibinfo{person}{Amit~Kumar Saha}, {and} \bibinfo{person}{David~B. Johnson}.}
  \bibinfo{year}{2003}\natexlab{}.
\newblock \bibinfo{title}{{CRAWDAD} dataset rice/ad\_hoc\_city (v.
  2003-09-11)}.
\newblock \bibinfo{howpublished}{Downloaded from
  \url{http://crawdad.org/rice/ad_hoc_city/20030911}}.   (\bibinfo{date}{Sept.}
  \bibinfo{year}{2003}).
\newblock
\showDOI{%
\url{http://dx.doi.org/10.15783/C73K5B}}


\bibitem[\protect\citeauthoryear{Krizhevsky, Sutskever, and Hinton}{Krizhevsky
  et~al\mbox{.}}{2012}]%
        {krizhevsky2012imagenet}
\bibfield{author}{\bibinfo{person}{Alex Krizhevsky}, \bibinfo{person}{Ilya
  Sutskever}, {and} \bibinfo{person}{Geoffrey~E Hinton}.}
  \bibinfo{year}{2012}\natexlab{}.
\newblock \showarticletitle{Imagenet classification with deep convolutional
  neural networks}. In \bibinfo{booktitle}{{\em Advances in neural information
  processing systems}}. \bibinfo{pages}{1097--1105}.
\newblock


\bibitem[\protect\citeauthoryear{LeCun, Bottou, Bengio, and Haffner}{LeCun
  et~al\mbox{.}}{1998}]%
        {lecun1998gradient}
\bibfield{author}{\bibinfo{person}{Yann LeCun}, \bibinfo{person}{L{\'e}on
  Bottou}, \bibinfo{person}{Yoshua Bengio}, {and} \bibinfo{person}{Patrick
  Haffner}.} \bibinfo{year}{1998}\natexlab{}.
\newblock \showarticletitle{Gradient-based learning applied to document
  recognition}.
\newblock \bibinfo{journal}{{\it Proc. IEEE}} \bibinfo{volume}{86},
  \bibinfo{number}{11} (\bibinfo{year}{1998}), \bibinfo{pages}{2278--2324}.
\newblock


\bibitem[\protect\citeauthoryear{Liaw and Wiener}{Liaw and Wiener}{2002}]%
        {liaw2002classification}
\bibfield{author}{\bibinfo{person}{Andy Liaw} {and} \bibinfo{person}{Matthew
  Wiener}.} \bibinfo{year}{2002}\natexlab{}.
\newblock \showarticletitle{Classification and regression by random forest}.
\newblock \bibinfo{journal}{{\em R news\/}} \bibinfo{volume}{2},
  \bibinfo{number}{3} (\bibinfo{year}{2002}), \bibinfo{pages}{18--22}.
\newblock


\bibitem[\protect\citeauthoryear{McBride}{McBride}{2015}]%
        {mcbride2015ridesourcing}
\bibfield{author}{\bibinfo{person}{Sean McBride}.}
  \bibinfo{year}{2015}\natexlab{}.
\newblock {\em \bibinfo{title}{Ridesourcing and the Taxi Marketplace}}.
\newblock \bibinfo{thesistype}{Ph.D. Dissertation}.
  \bibinfo{school}{Dissertation thesis submitted to Boston College, College of
  Arts and Sciences}.
\newblock


\bibitem[\protect\citeauthoryear{Morning}{Morning}{2015}]%
        {Company_car_didi}
\bibfield{author}{\bibinfo{person}{Shanghai Morning}.}
  \bibinfo{year}{2015}\natexlab{}.
\newblock \bibinfo{title}{{Company-owned cars for Didi services (in Chinese).}}
\newblock
  \bibinfo{howpublished}{\url{http://newspaper.jfdaily.com/xwcb/html/2015-08/05/content_119186.htm}}.
    (\bibinfo{year}{2015}).
\newblock
\newblock
\shownote{Accessed: 2017-01-15.}


\bibitem[\protect\citeauthoryear{MSRA}{MSRA}{2017}]%
        {beijingTaxi}
\bibfield{author}{\bibinfo{person}{MSRA}.} \bibinfo{year}{2017}\natexlab{}.
\newblock \bibinfo{title}{Beijing Taxi Data}.
\newblock
  \bibinfo{howpublished}{\url{https://www.microsoft.com/en-us/research/publication/t-drive-trajectory-data-sample/}}.
    (\bibinfo{year}{2017}).
\newblock
\newblock
\shownote{Accessed: 2017-01-15.}


\bibitem[\protect\citeauthoryear{News}{News}{2015}]%
        {NYC_camera}
\bibfield{author}{\bibinfo{person}{Daily News}.}
  \bibinfo{year}{2015}\natexlab{}.
\newblock \bibinfo{title}{{Expand New York City's surveillance camera
  network.}}
\newblock
  \bibinfo{howpublished}{\url{http://www.nydailynews.com/opinion/bryan-schonfeld-expand-nyc-surveillance-camera-network-article-1.2117122}}.
    (\bibinfo{year}{2015}).
\newblock
\newblock
\shownote{Accessed: 2017-01-15.}


\bibitem[\protect\citeauthoryear{Norris and Armstrong}{Norris and
  Armstrong}{1999}]%
        {norris1999maximum}
\bibfield{author}{\bibinfo{person}{Clive Norris} {and} \bibinfo{person}{Gary
  Armstrong}.} \bibinfo{year}{1999}\natexlab{}.
\newblock \bibinfo{booktitle}{{\em The maximum surveillance society: The rise
  of CCTV}}.
\newblock \bibinfo{publisher}{Berg Publishers}.
\newblock


\bibitem[\protect\citeauthoryear{Ondrej, Zboril~Frantisek, and Martin}{Ondrej
  et~al\mbox{.}}{2007}]%
        {ondrej2007algorithmic}
\bibfield{author}{\bibinfo{person}{Martinsk{\`y} Ondrej}, \bibinfo{person}{V
  Zboril~Frantisek}, {and} \bibinfo{person}{Drahansk{\`y} Martin}.}
  \bibinfo{year}{2007}\natexlab{}.
\newblock \showarticletitle{Algorithmic and mathematical principles of
  automatic number plate recognition systems}.
\newblock \bibinfo{journal}{{\em BRNO University of technology\/}}
  (\bibinfo{year}{2007}), \bibinfo{pages}{10}.
\newblock


\bibitem[\protect\citeauthoryear{Pan and Yang}{Pan and Yang}{2010}]%
        {pan2010survey}
\bibfield{author}{\bibinfo{person}{Sinno~Jialin Pan} {and}
  \bibinfo{person}{Qiang Yang}.} \bibinfo{year}{2010}\natexlab{}.
\newblock \showarticletitle{A survey on transfer learning}.
\newblock \bibinfo{journal}{{\em IEEE Transactions on knowledge and data
  engineering\/}} \bibinfo{volume}{22}, \bibinfo{number}{10}
  (\bibinfo{year}{2010}), \bibinfo{pages}{1345--1359}.
\newblock


\bibitem[\protect\citeauthoryear{Piorkowski, Sarafijanovic-Djukic, and
  Grossglauser}{Piorkowski et~al\mbox{.}}{2009}]%
        {SFTaxi}
\bibfield{author}{\bibinfo{person}{Michal Piorkowski}, \bibinfo{person}{Natasa
  Sarafijanovic-Djukic}, {and} \bibinfo{person}{Matthias Grossglauser}.}
  \bibinfo{year}{2009}\natexlab{}.
\newblock \bibinfo{title}{{CRAWDAD} dataset epfl/mobility (v. 2009-02-24)}.
\newblock \bibinfo{howpublished}{Downloaded from
  \url{http://crawdad.org/epfl/mobility/20090224}}.   (\bibinfo{date}{Feb.}
  \bibinfo{year}{2009}).
\newblock
\showDOI{%
\url{http://dx.doi.org/10.15783/C7J010}}


\bibitem[\protect\citeauthoryear{Rayle, Dai, Chan, Cervero, and Shaheen}{Rayle
  et~al\mbox{.}}{2016}]%
        {rayle2016just}
\bibfield{author}{\bibinfo{person}{Lisa Rayle}, \bibinfo{person}{Danielle Dai},
  \bibinfo{person}{Nelson Chan}, \bibinfo{person}{Robert Cervero}, {and}
  \bibinfo{person}{Susan Shaheen}.} \bibinfo{year}{2016}\natexlab{}.
\newblock \showarticletitle{Just a better taxi? A survey-based comparison of
  taxis, transit, and ridesourcing services in San Francisco}.
\newblock \bibinfo{journal}{{\em Transport Policy\/}}  \bibinfo{volume}{45}
  (\bibinfo{year}{2016}), \bibinfo{pages}{168--178}.
\newblock


\bibitem[\protect\citeauthoryear{Rothberg}{Rothberg}{2015}]%
        {uber_under_table}
\bibfield{author}{\bibinfo{person}{Daniel Rothberg}.}
  \bibinfo{year}{2015}\natexlab{}.
\newblock \bibinfo{title}{{Uber driver suspended after caught giving ride
  outside app.}}
\newblock
  \bibinfo{howpublished}{\url{https://lasvegassun.com/news/2015/oct/06/uber-driver-suspended-after-he-was-caught-giving-a/}}.
    (\bibinfo{year}{2015}).
\newblock
\newblock
\shownote{Accessed: 2017-02-15.}


\bibitem[\protect\citeauthoryear{Shaheen, Cohen, and Zohdy}{Shaheen
  et~al\mbox{.}}{2016}]%
        {sharedMobility}
\bibfield{author}{\bibinfo{person}{Susan Shaheen}, \bibinfo{person}{Adam
  Cohen}, {and} \bibinfo{person}{Ismail Zohdy}.}
  \bibinfo{year}{2016}\natexlab{}.
\newblock \showarticletitle{Shared Mobility: Current Practices and Guiding
  Principles}.
\newblock \bibinfo{journal}{{\em U.S. Department of Transportation Federal
  Highway Administration Technical Report No. FHWA-HOP-16-022\/}}
  (\bibinfo{year}{2016}).
\newblock


\bibitem[\protect\citeauthoryear{SJTU}{SJTU}{2017}]%
        {shanghaiTaxi}
\bibfield{author}{\bibinfo{person}{SJTU}.} \bibinfo{year}{2017}\natexlab{}.
\newblock \bibinfo{title}{SUVnet-Trace Data}.
\newblock
  \bibinfo{howpublished}{\url{http://wirelesslab.sjtu.edu.cn/taxi_trace_data.html}}.
    (\bibinfo{year}{2017}).
\newblock
\newblock
\shownote{Accessed: 2017-01-15.}


\bibitem[\protect\citeauthoryear{Srivastava, Hinton, Krizhevsky, Sutskever, and
  Salakhutdinov}{Srivastava et~al\mbox{.}}{2014}]%
        {srivastava2014dropout}
\bibfield{author}{\bibinfo{person}{Nitish Srivastava},
  \bibinfo{person}{Geoffrey~E Hinton}, \bibinfo{person}{Alex Krizhevsky},
  \bibinfo{person}{Ilya Sutskever}, {and} \bibinfo{person}{Ruslan
  Salakhutdinov}.} \bibinfo{year}{2014}\natexlab{}.
\newblock \showarticletitle{Dropout: a simple way to prevent neural networks
  from overfitting.}
\newblock \bibinfo{journal}{{\em Journal of Machine Learning Research\/}}
  \bibinfo{volume}{15}, \bibinfo{number}{1} (\bibinfo{year}{2014}),
  \bibinfo{pages}{1929--1958}.
\newblock


\bibitem[\protect\citeauthoryear{Sun and Ban}{Sun and Ban}{2013}]%
        {sun2013vehicle}
\bibfield{author}{\bibinfo{person}{Zhanbo Sun} {and}
  \bibinfo{person}{Xuegang~Jeff Ban}.} \bibinfo{year}{2013}\natexlab{}.
\newblock \showarticletitle{Vehicle classification using GPS data}.
\newblock \bibinfo{journal}{{\em Transportation Research Part C: Emerging
  Technologies\/}}  \bibinfo{volume}{37} (\bibinfo{year}{2013}),
  \bibinfo{pages}{102--117}.
\newblock


\bibitem[\protect\citeauthoryear{TLC}{TLC}{2017}]%
        {NYTaxi}
\bibfield{author}{\bibinfo{person}{NYC TLC}.} \bibinfo{year}{2017}\natexlab{}.
\newblock \bibinfo{title}{NYC Taxi Data}.
\newblock
  \bibinfo{howpublished}{\url{http://www.nyc.gov/html/tlc/html/about/trip_record_data.shtml}}.
    (\bibinfo{year}{2017}).
\newblock
\newblock
\shownote{Accessed: 2017-01-15.}


\bibitem[\protect\citeauthoryear{Wang, Yuan, Lian, Xu, Xie, Chen, and Rui}{Wang
  et~al\mbox{.}}{2015}]%
        {wang2015regularity}
\bibfield{author}{\bibinfo{person}{Yingzi Wang}, \bibinfo{person}{Nicholas~Jing
  Yuan}, \bibinfo{person}{Defu Lian}, \bibinfo{person}{Linli Xu},
  \bibinfo{person}{Xing Xie}, \bibinfo{person}{Enhong Chen}, {and}
  \bibinfo{person}{Yong Rui}.} \bibinfo{year}{2015}\natexlab{}.
\newblock \showarticletitle{Regularity and conformity: location prediction
  using heterogeneous mobility data}. In \bibinfo{booktitle}{{\em Proceedings
  of KDD}}. ACM, \bibinfo{pages}{1275--1284}.
\newblock


\bibitem[\protect\citeauthoryear{Wang, Zheng, and Xue}{Wang
  et~al\mbox{.}}{2014}]%
        {wang2014travel}
\bibfield{author}{\bibinfo{person}{Yilun Wang}, \bibinfo{person}{Yu Zheng},
  {and} \bibinfo{person}{Yexiang Xue}.} \bibinfo{year}{2014}\natexlab{}.
\newblock \showarticletitle{Travel time estimation of a path using sparse
  trajectories}. In \bibinfo{booktitle}{{\em Proceedings of KDD}}. ACM,
  \bibinfo{pages}{25--34}.
\newblock


\bibitem[\protect\citeauthoryear{Wikipedia}{Wikipedia}{2017}]%
        {wiki_uber_legal}
\bibfield{author}{\bibinfo{person}{Wikipedia}.}
  \bibinfo{year}{2017}\natexlab{}.
\newblock \bibinfo{title}{{Uber protests and legal actions.}}
\newblock
  \bibinfo{howpublished}{\url{https://en.wikipedia.org/wiki/Uber_protests_and_legal_actions}}.
    (\bibinfo{year}{2017}).
\newblock
\newblock
\shownote{Accessed: 2017-02-17.}


\bibitem[\protect\citeauthoryear{Yang and Maxwell}{Yang and Maxwell}{2011}]%
        {yang2011information}
\bibfield{author}{\bibinfo{person}{Tung-Mou Yang} {and}
  \bibinfo{person}{Terrence~A Maxwell}.} \bibinfo{year}{2011}\natexlab{}.
\newblock \showarticletitle{Information-sharing in public organizations: A
  literature review of interpersonal, intra-organizational and
  inter-organizational success factors}.
\newblock \bibinfo{journal}{{\em Government Information Quarterly\/}}
  \bibinfo{volume}{28}, \bibinfo{number}{2} (\bibinfo{year}{2011}),
  \bibinfo{pages}{164--175}.
\newblock


\bibitem[\protect\citeauthoryear{Yuen, King, and Leung}{Yuen
  et~al\mbox{.}}{2011}]%
        {yuen2011survey}
\bibfield{author}{\bibinfo{person}{Man-Ching Yuen}, \bibinfo{person}{Irwin
  King}, {and} \bibinfo{person}{Kwong-Sak Leung}.}
  \bibinfo{year}{2011}\natexlab{}.
\newblock \showarticletitle{A survey of crowdsourcing systems}. In
  \bibinfo{booktitle}{{\em SocialCom}}. IEEE, \bibinfo{pages}{766--773}.
\newblock


\bibitem[\protect\citeauthoryear{Zhang, Li, Zhou, Chen, Sun, and Li}{Zhang
  et~al\mbox{.}}{2011}]%
        {zhang2011ibat}
\bibfield{author}{\bibinfo{person}{Daqing Zhang}, \bibinfo{person}{Nan Li},
  \bibinfo{person}{Zhi-Hua Zhou}, \bibinfo{person}{Chao Chen},
  \bibinfo{person}{Lin Sun}, {and} \bibinfo{person}{Shijian Li}.}
  \bibinfo{year}{2011}\natexlab{}.
\newblock \showarticletitle{iBAT: detecting anomalous taxi trajectories from
  GPS traces}. In \bibinfo{booktitle}{{\em Proceedings of the 13th
  international conference on Ubiquitous computing}}. ACM,
  \bibinfo{pages}{99--108}.
\newblock


\bibitem[\protect\citeauthoryear{Zhang, Zheng, and Qi}{Zhang
  et~al\mbox{.}}{2017}]%
        {zhang2016deep}
\bibfield{author}{\bibinfo{person}{Junbo Zhang}, \bibinfo{person}{Yu Zheng},
  {and} \bibinfo{person}{Dekang Qi}.} \bibinfo{year}{2017}\natexlab{}.
\newblock \showarticletitle{Deep Spatio-Temporal Residual Networks for Citywide
  Crowd Flows Prediction}. In \bibinfo{booktitle}{{\em AAAI'17}}.
\newblock


\bibitem[\protect\citeauthoryear{Zheng}{Zheng}{2015}]%
        {zheng2015trajectory}
\bibfield{author}{\bibinfo{person}{Yu Zheng}.} \bibinfo{year}{2015}\natexlab{}.
\newblock \showarticletitle{Trajectory data mining: an overview}.
\newblock \bibinfo{journal}{{\em ACM Transactions on Intelligent Systems and
  Technology (TIST)\/}} \bibinfo{volume}{6}, \bibinfo{number}{3}
  (\bibinfo{year}{2015}), \bibinfo{pages}{29}.
\newblock


\bibitem[\protect\citeauthoryear{Zheng, Chen, Li, Xie, and Ma}{Zheng
  et~al\mbox{.}}{2010}]%
        {zheng2010understanding}
\bibfield{author}{\bibinfo{person}{Yu Zheng}, \bibinfo{person}{Yukun Chen},
  \bibinfo{person}{Quannan Li}, \bibinfo{person}{Xing Xie}, {and}
  \bibinfo{person}{Wei-Ying Ma}.} \bibinfo{year}{2010}\natexlab{}.
\newblock \showarticletitle{Understanding transportation modes based on GPS
  data for web applications}.
\newblock \bibinfo{journal}{{\em ACM Transactions on the Web (TWEB)\/}}
  \bibinfo{volume}{4}, \bibinfo{number}{1} (\bibinfo{year}{2010}),
  \bibinfo{pages}{1}.
\newblock


\bibitem[\protect\citeauthoryear{Zheng, Liu, and Hsieh}{Zheng
  et~al\mbox{.}}{2013}]%
        {zheng2013u}
\bibfield{author}{\bibinfo{person}{Yu Zheng}, \bibinfo{person}{Furui Liu},
  {and} \bibinfo{person}{Hsun-Ping Hsieh}.} \bibinfo{year}{2013}\natexlab{}.
\newblock \showarticletitle{U-Air: when urban air quality inference meets big
  data}. In \bibinfo{booktitle}{{\em Proceedings of KDD}}. ACM,
  \bibinfo{pages}{1436--1444}.
\newblock


\bibitem[\protect\citeauthoryear{Zhou}{Zhou}{2012}]%
        {zhou2012ensemble}
\bibfield{author}{\bibinfo{person}{Zhi-Hua Zhou}.}
  \bibinfo{year}{2012}\natexlab{}.
\newblock \bibinfo{booktitle}{{\em Ensemble methods: foundations and
  algorithms}}.
\newblock \bibinfo{publisher}{CRC press}.
\newblock


\end{thebibliography}
